\documentclass[jair,twoside,11pt]{article}
\usepackage{jair,rawfonts}
\usepackage{apacite}

\jairheading{68}{2020}{503--540}{09/2019}{07/2020}
\ShortHeadings{Ontology Reasoning with Deep Neural Networks}{Hohenecker and Lukasiewicz}
\firstpageno{503}

\usepackage[nolist]{acronym}
\usepackage[linesnumbered,lined,boxed,commentsnumbered]{algorithm2e}
\usepackage{amsfonts}
\usepackage{amsmath}
\usepackage{array}
\usepackage[english]{babel}
\usepackage{color}
\usepackage{colortbl}
\usepackage[autostyle]{csquotes}  
\usepackage{dsfont}               
\usepackage[T1]{fontenc}          
\usepackage{enumitem}
\usepackage{ifthen}
\usepackage{lineno}
\usepackage{listings}
\usepackage{mathtools}
\usepackage{multirow}
\usepackage{natbib}
\usepackage{soul}
\usepackage{tikz}
\usepackage[normalem]{ulem}
\usepackage{xcolor}

\usetikzlibrary{
    arrows,
    arrows.meta,
    fadings,
    patterns,
    positioning,
    shapes.geometric
}

\makeatletter
\newcommand{\tinytiny}{\@setfontsize{\srcsize}{5pt}{5pt}}
\renewenvironment*{figure}{\@float{figure}}{\end@float}
\makeatother
\SetAlCapSty{}      
\SetAlCapSkip{1em}  

\input{toy-dataset-figure.tex}
\definecolor{highlight2}{RGB}{0, 245, 255}

\newcommand{\BgColor}         {black!80}

\newcommand{\EmbOffsetX}      {20}
\newcommand{\EmbOffsetY}      {-6}
\newcommand{\EmbeddingDim}    {10}
\newcommand{\HighlightPadding}{0.15}
\newcommand{\NumEmbeddings}   {24}
\newcommand{\VecElemSize}     {0.5}

\newcommand{\ModelFigure}{%

    \begin{figure}[t]
        \centering
        \renewcommand{\baselinestretch}{1}
        \setlength{\baselineskip}{0mm}
        
        \begin{tikzpicture}[
            scale = 0.485,
            data/.style = {
                rectangle,
                draw,
                fill,
                white,
                text = black,
                font = \tiny\sf,
                rounded corners = 3
            },
            EdgeLabel/.style = {
                font = \tiny\sf
            },
            Embedding/.style = {
                rectangle,
                draw,
                font = \tiny\sf,
                line width = 1,
                text = white,
                pattern = north west lines,
                minimum width = 50pt
            },
            InputEdge/.style = {
                white,
                line width = 1
            },
            Index/.style = {
                font = \tiny\sf,
                text width = 30pt,
                align = left
            },
            LookUp/.style = {
                circle,
                draw = white,
                text = white,
                font = \tiny\sf,
                align = center,
                rounded corners = 3,
                inner sep = 1pt
            },
            LookUpEdge/.style = {
                white,
                line width = 1
            },
            MyArrow/.style = {
                -{Latex[length=3mm, width=1.8mm]}
            },
            Section/.style = {
                fill,
                black!10,
                rounded corners = 2
            },
            SectionLabel/.style = {
                anchor = north,
                font = \scriptsize\sf
            },
            SectionSeparator/.style = {
                line width = 1,
                black!30,
                dotted
            },
            TimeEdge/.style = {
                MyArrow,
                orange!90,
                line width = 1
            },
            Triple/.style = {
                rectangle,
                draw,
                fill,
                text = black,
                font = \tiny\sf,
                rounded corners = 10pt,
                minimum height = 20pt
            },
            UpdateLayer/.style = {
                circle,
                draw,
                fill,
                white,
                inner sep = 0,
                minimum width = 80pt,
                align = center
            }
        ]
        
            \draw[SectionSeparator] (1.6,  9) -- +(0, -18);
            \draw[SectionSeparator] (18.5, 9) -- +(0, -18);
            \draw[SectionSeparator] (-4.8, -8) -- (26.4, -8);
            
            \draw[Section, black!30] (-4.8, 8) rectangle +( 6.3, 1);
            \draw[Section, black!20] ( 1.7, 8) rectangle +(16.7, 1);
            \draw[Section, black!10] (18.6, 8) rectangle +( 7.8, 1);
            
            \node[SectionLabel] at (-1.65,  9.0) {Data};
            \node[SectionLabel] at (10.05, 9.0) {Model};
            \node[SectionLabel] at (22.5,  9.0) {Embeddings};
            
            \node[SectionLabel, anchor = north] at (-1.65,  -8.1) {\textbf{a}};
            \node[SectionLabel, anchor = north] at (10.05,  -8.1) {\textbf{b}};
            \node[SectionLabel, anchor = north] at (22.5,  -8.1) {\textbf{c}};

            \draw[fill, \BgColor] (-3, 0) arc (180:90:2)
                                          --  (1, 2)
                                          arc (270:360:1)
                                          --  (2, -3)
                                          arc (0:90:1)
                                          --  (-1, -2)
                                          arc (270:180:2);
            \draw[fill, white] (-2, 0) arc (180:90:1)
                                      --  ( 2, 1)
                                      --  ( 2, -1)
                                      --  (-1, -1)
                                      arc (270:180:1);
            \draw[fill, \BgColor, rounded corners = 20] (2, 6) rectangle (17, -6);
            \node[circle, draw, fill, \BgColor, inner sep = 0, minimum width = 120pt] at (14, 0) {};
            
            \draw[SectionSeparator] (1.6,  1) -- +(0, -2);

            \node[Index] at (-3, -4) {\#$k\! -\! 1$};
            \node[Triple, draw = orange, fill = orange] (last)    at ( 0, -4) {\ $\langle s, p, o \rangle$\ \ };
            \node[Index] at (-3,  0) {\#$k$};
            \node[Triple, draw = orange!80, fill = orange!80] (current) at ( 0,  0) {\ $\langle s, p, o \rangle$\ \ };
            \node[Index] at (-3,  4) {\#$k\! +\! 1$};
            \node[Triple, draw = orange!60, fill = orange!60] (next)    at ( 0,  4) {\ $\langle s, p, o \rangle$\ \ };
            
            \draw[TimeEdge, orange!60] (next) -- (0, 2);
            \draw[TimeEdge, orange] (0, -2.015) -- (last);

            \draw[line width = 0.5, dashed, black!20] (-4.5, -4) arc (180:90:0.7)  -- +(3, 0);
            \draw[line width = 0.5, dashed, black!20] (-4.5, -4) arc (180:270:0.7) -- +(3, 0);
            
            \draw[line width = 0.5, dashed, black!20] (-4.5, 0) arc (180:90:0.7)  -- +(1, 0);
            \draw[line width = 0.5, dashed, black!20] (-4.5, 0) arc (180:270:0.7) -- +(1, 0);
            \draw[line width = 0.5, dashed, black!20] (-1.6,  0.7) -- +(0.8, 0);
            \draw[line width = 0.5, dashed, black!20] (-1.6, -0.7) -- +(0.8, 0);
            
            \draw[line width = 0.5, dashed, black!20] (-4.5, 4) arc (180:90:0.7)  -- +(3, 0);
            \draw[line width = 0.5, dashed, black!20] (-4.5, 4) arc (180:270:0.7) -- +(3, 0);

            \node[circle, draw, fill, white, inner sep = 0pt, minimum width = 4] (infork) at (3.25, 0) {};
            
            \node[data] (subject)   at (6,  4) {subject};
            \node[data] (predicate) at (6,  0) {predicate};
            \node[data] (object)    at (6, -4) {object};
            
            \node[LookUp] (subLook)  at (9.5,  4) {look\\up};
            \node[LookUp] (objLook)  at (9.5, -4) {look\\up};
            
            \node[Embedding, draw = green, pattern color = green]           (subEmb) at (14,  4) {};
            \node[Embedding, draw = highlight2, pattern color = highlight2] (objEmb) at (14, -4) {};
            
            \node[UpdateLayer] (model) at (14, 0) {%
                \begin{tikzpicture}[
                    xscale = 0.6,
                    yscale = 0.8,
                    Edge/.style = {
                        black,
                        line width = 1,
                        -{Latex[length=2mm, width=1mm]}
                    },
                    Vertex/.style = {
                        circle,
                        draw,
                        fill,
                        black,
                        inner sep = 0,
                        minimum width = 7pt
                    }
                ]
                    \node[Vertex, white] at (1.5, 0) {};
                    
                    \node[Vertex] (v00) at (0, 0.2) {};
                    \node[Vertex] (v01) at (1, 0.2) {};
                    \node[Vertex] (v02) at (2, 0.2) {};
                    \node[Vertex] (v03) at (3, 0.2) {};
                    \node[Vertex] (v10) at (0.5, 1) {};
                    \node[Vertex] (v11) at (2.5, 1) {};
                    \node[Vertex] (v20) at (1.5, 1.5) {};
                    
                    \draw[Edge] (v00) -- (v10);
                    \draw[Edge] (v01) -- (v10);
                    \draw[Edge] (v02) -- (v11);
                    \draw[Edge] (v03) -- (v11);
                    \draw[Edge] (v10) -- (v20);
                    \draw[Edge] (v11) -- (v20);
                \end{tikzpicture}%
            };
            
            \draw[white, line width = 1] (1.5, 0) -- (infork);
            \draw[white, line width = 1] (infork) -- +(0, 3)
                                                  arc (180:90:1)
                                                  [MyArrow]-- (subject);
            \draw[InputEdge, MyArrow] (infork) -- (predicate);
            \draw[white, line width = 1] (infork) -- +(0, -3)
                                                  arc (180:270:1)
                                                  [MyArrow]-- (object);

            \draw[InputEdge, MyArrow] (subject)   -- (subLook);
            \draw[InputEdge, MyArrow] (predicate) -- (model);
            \draw[InputEdge, MyArrow] (object)    -- (objLook);
            
            \draw[InputEdge, MyArrow, green]      (subEmb) -- (model);
            \draw[InputEdge, MyArrow, highlight2] (objEmb) -- (model);

            \pgfmathsetseed{123}
            \foreach \x in {1,...,\EmbeddingDim}{
			    \foreach \y in {1,...,\NumEmbeddings}{
                    \ifthenelse{\y = 4 \OR \y = 16}
                    {%
                        \pgfmathparse{0.4 + rnd * 0.4}
                        \edef\Offset{\pgfmathresult}
                        
                        \ifthenelse{\y = 4}
                        {%
                            \edef\RedValue{0}
                            \edef\GreenValue{0.961}
                            \edef\BlueValue{1}
                        }
                        {%
                            \edef\RedValue{0}
                            \edef\GreenValue{1}
                            \edef\BlueValue{0}
                        }
                    }
                    {%
                        \pgfmathparse{0.4 + rnd * 0.4}
                        \edef\RedValue{\pgfmathresult}
                        \edef\GreenValue{\pgfmathresult}
                        \edef\BlueValue{\pgfmathresult}
                    }
                    
                    \definecolor{RandomGray}{rgb}{\RedValue , \GreenValue, \BlueValue}

                    \draw[fill, RandomGray, line width = 0.1, draw = white]
                        (
                            \EmbOffsetX + \x * \VecElemSize - \VecElemSize,
                            \EmbOffsetY + \y * \VecElemSize - \VecElemSize
                        )
                        rectangle
                        +(\VecElemSize, \VecElemSize);
                }
            }
            
            \draw[draw, highlight2, line width = 1]
                (
                    \EmbOffsetX - \HighlightPadding,
                    \EmbOffsetY + 3 * \VecElemSize - \HighlightPadding
                )
                rectangle
                +(
                    \EmbeddingDim * \VecElemSize + 2 * \HighlightPadding,
                    \VecElemSize + 2 * \HighlightPadding
                );
            \draw[draw, green, line width = 1]
                (
                    \EmbOffsetX - \HighlightPadding,
                    \EmbOffsetY + 15 * \VecElemSize - \HighlightPadding
                )
                rectangle
                +(
                    \EmbeddingDim * \VecElemSize + 2 * \HighlightPadding,
                    \VecElemSize + 2 * \HighlightPadding
                );

            \draw[LookUpEdge, green] (25.15, 1.75) -- (26, 1.75)  
                                                   -- (26, 7)
                                                   -- (9.5, 7) 
                                                   -- (9.5, 4.8)
                                                   arc (90:0:0.8)
                                                   [MyArrow]-- (subEmb);
            \draw[LookUpEdge, highlight2] (25.15, -4.25) -- (26, -4.25)  
                                                         -- (26, -7)
                                                         -- (9.5, -7) 
                                                         -- (9.5, -4.8)
                                                        arc (270:360:0.8)
                                                        [MyArrow]-- (objEmb);
            
            \draw[white, line width = 9] (16.8, 0) -- (18.3, 0);
            \draw[LookUpEdge, green] (16.83, 0.1) -- (19, 0.1)
                                                  -- (19, 1.75)
                                                  [MyArrow]-- (19.85, 1.75);
            \draw[LookUpEdge, highlight2] (16.83, -0.1) -- (19, -0.1)
                                                  -- (19, -4.25)
                                                  [MyArrow]-- (19.85, -4.25);
        
        \end{tikzpicture}
    
        \caption{%
            \textbf{(a)} To generate embeddings for the individuals in a knowledge base, the
                         model iterates over all its facts.
                         To that end, it considers one triple at a time in arbitrary order, and,
                         depending on the respective dataset, might repeat this process several
                         times.
            \textbf{(b)} Whenever a fact is read, the model fetches the embeddings of the
                         individuals that appear in the according triple, and feeds them into an
                         update layer for the respective predicate.
                         This layer yields updated versions of the embeddings that have been
                         provided, which are then stored in place of the previous ones.
                         The figure illustrates such an update for the case that a triple
                         describes a relation between two individuals.
                         In contrast to this, updates based on facts specifying class memberships
                         involve a single individual embedding only.
            \textbf{(c)} Starting from randomly generated vectors, embeddings are updated step by
                         step in order to encode both facts and inferences about the individuals
                         that they represent.
        }
        \label{fig:embedding-generation}
    \end{figure}

}

\input{embeddings-figure.tex}

\newcommand{\MyQuote}[1]{\frqq\textit{#1}\flqq}
\newcommand{\RowColor}{gray!10}

\newcommand{\ResultTable}[4]
{
    {
        \tiny
        \renewcommand{\arraystretch}{1}
    
        \begin{tabular}{| l l | c c c c c c |}
        
            \hline
            \multicolumn{8}{| c |}{\cellcolor{black}\color{white}\textbf{#1}} \\
            \hline\hline
            &
            \textbf{name} &
                \textbf{F1 score} &
                \textbf{AUC-PR} &
                \textbf{total} &
                \textbf{accuracy on} &
                \textbf{accuracy on} &
                \textbf{\# of triples} \\
            &
            &
                &
                &
                \textbf{accuracy} &
                \textbf{positives} &
                \textbf{negatives} &
                \textbf{(pos./neg.)} \\
            \hline
            #2
            \hline\hline
            \multicolumn{2}{| c |}{\textbf{TOTAL}} &
                \multicolumn{2}{c |}{\textbf{F1 score}} &
                \multicolumn{2}{c |}{\textbf{AUC-PR}} &
                \multicolumn{2}{c |}{\textbf{accuracy}}  \\
            \hline
            \multicolumn{2}{| c |}{} &
            #3
            \hline
        \end{tabular}
    }
}

\newcommand{\BigO}[1]{\mathcal{O} ( #1 ) }                
\newcommand{\Mat}[1]{\mathbf{#1}}                         
\DeclarePairedDelimiterX{\Norm}[1]{\lVert}{\rVert_2}{#1}  

\newcommand{\Prob}[1]{\mathbb{P} \left\{ #1 \right\}}     
\newcommand{\R}{\mathbb{R}}                               
\newcommand{\UniformDist}[2]{\mathcal{U} ( #1, #2 )}      
\newcommand{\Vect}[1]{\mathbf{#1}}                        

\newcommand{\Candidate}[1]{\hat{\Vect{e}}_{#1}}           
\newcommand{\Class}[1]{C_{#1}}
\newcommand{\ClassSymbol}[0]{C}
\newcommand{\Classes}[0]{\text{\it classes}}
\newcommand{\Embedding}[1]{\Vect{e}_{#1}}                 
\newcommand{\EmbeddingSize}[0]{d}                         
\newcommand{\Facts}[0]{D}                                 
\newcommand{\FullUpdate}[3]{%
    \text{\it Update} ( \Triple{#1}{#2}{#3} )%
}
\newcommand{\GateSourceUpdate}[3]{
    \Vect{g}^{\triangleleft #2} ( #1 , #3 )%
}
\newcommand{\GateSourceUpdateWeightsSource}[1]{
    \Mat{V}_1^{\triangleleft #1}%
}
\newcommand{\GateSourceUpdateWeightsTarget}[1]{
    \Mat{V}_2^{\triangleleft #1}%
}
\newcommand{\Incidence}[1]{\mathds{1}_{#1}}               
\newcommand{\Individuals}[0]{\text{\it individuals}}

\newcommand{\InfRelations}[1][]{\text{\it Rel}^{\,\Entails}\ifthenelse{\equal{#1}{}}{}{( #1 )}}

\newcommand{\Member}[0]{\text{\tt MemberOf}}
\newcommand{\NumClasses}[0]{m}                            
\newcommand{\NumRelations}[0]{n}                          
\newcommand{\OKB}[0]{\mathit{K\!B}}                       
\newcommand{\Ontology}[0]{\Sigma}                         

\newcommand{\ReLU}[0]{\text{\it ReLU}}                    

\newcommand{\RelationSymbol}[0]{R}
\newcommand{\RRN}[1]{\text{\it RRN}_{#1}}
\newcommand{\Triple}[3]{\langle #1, #2, #3 \rangle}       
\newcommand{\UpdateSource}[3]{
    \text{\it Update}^{\triangleleft #2} ( #1, #3 )%
}
\newcommand{\UpdateTarget}[3]{
    \text{\it Update}^{#2 \triangleright} ( #1, #3 )%
}
\newcommand{\UpdateWeightsSource}[1]{\Mat{W}_1^{#1}}      
\newcommand{\UpdateWeightsTarget}[1]{\Mat{W}_2^{#1}}      
\newcommand{\UpdateWeightsInteraction}[1]{\Vect{w}^{#1}}  

\newcommand{\MlpClasses}{\text{\textit{MLP}}^\text{(classes)}}  
\newcommand{\MlpRelations}[1]{\text{\textit{MLP}}^{(#1)}}       

\newcommand{\Parameters}{\boldsymbol\theta}  

\newcommand{\BodyVar}[1]{\beta_{#1}}           
\newcommand{\Constants}{\Delta}                
\newcommand{\Constraint}{\gamma}               
\newcommand{\Entails}{{\mid}\!{\sim}}          
\newcommand{\HeadVar}{\alpha}                  
\newcommand{\Interpretation}{I}                
\newcommand{\Schema}{\mathcal{R}}              
\newcommand{\Variables}{\mathcal{V}}           

\newcommand{\object}{$\mathit{object}$}
\newcommand{\predicate}{$\mathit{predicate}$}
\newcommand{\subject}{$\mathit{subject}$}

\begin{document}


\title{Ontology Reasoning with Deep Neural Networks}

\author{%
    \name  Patrick Hohenecker
    \email patrick.hohenecker@cs.ox.ac.uk       \\
    \addr  Department of Computer Science       \\
           University of Oxford, UK             \\
    \AND
    \name  Thomas Lukasiewicz
    \email thomas.lukasiewicz@cs.ox.ac.uk       \\
    \addr  Department of Computer Science       \\
           University of Oxford, UK
}

\maketitle



\begin{abstract}
    The ability to conduct logical reasoning is a fundamental aspect of intelligent human behavior,
    and thus an important problem along the way to human-level artificial intelligence.
    Traditionally, logic-based symbolic methods from the field of knowledge representation and
    reasoning have been used to equip agents with capabilities that resemble human logical reasoning
    qualities.
    More recently, however, there has been an increasing interest in using machine learning rather
    than logic-based symbolic formalisms to tackle these tasks.
    In this paper, we employ state-of-the-art methods for training deep neural networks to devise a
    novel model that is able to learn how to effectively perform logical reasoning in the form of
    basic ontology reasoning.
    This is an important and at the same time very natural logical reasoning task, which is why the
    presented approach is applicable to a plethora of important real-world problems.
    We present the outcomes of several experiments, which show that our model is able to learn to
    perform highly accurate ontology reasoning on very large, diverse, and challenging benchmarks.
    Furthermore, it turned out that the suggested approach suffers much less from different
    obstacles that prohibit logic-based symbolic reasoning, and, at the same time, is surprisingly
    plausible from a biological point of view.
\end{abstract}


\section{Introduction}

Implementing human-like logical reasoning has been among the major goals of artificial intelligence
research ever since, and has recently also enjoyed increasing attention in the field of machine
learning.
However, a noticeable commonality of previous approaches in this area is that they, with a few very
recent exceptions
\citep{Serafini2016,Cai2017,Rocktaschel2017,Cingillioglu2018,Dai2018,Evans2018,Manhaeve2018},
entertain a quite informal notion of reasoning, which is often simply identified with a particular
kind of prediction task.
This contrasts the (traditional) understanding of reasoning as an application of mathematical proof
theory, like it is used in the context of logic-based knowledge representation and reasoning~(KRR).
Interestingly, however, it can be observed that, under certain provisions, even the best
reasoning models based on machine learning are still not in a position to compete with their
symbolic counterparts.
To close this gap between learning-based and KRR methods, we develop a novel model architecture,
called \emph{recursive reasoning network} (RRN), which makes use of recent advances in the area of
deep neural networks \citep{Bengio2009}.
By design, this model is much closer to logic-based symbolic methods than most of the other
learning-based approaches, but the fact that it employs machine learning allows for overcoming many
of the obstacles that we encounter with KRR methods in practice.
Furthermore, while it might seem paradoxical to elect logic-based symbolic methods as a starting
point for implementing human-like logical reasoning, blending those concepts with deep learning
leads to compelling results, just as it opens up interesting new perspectives on the considered
problem. 

Articles on reasoning based on machine learning commonly assume a particular application, such as
reasoning about natural language \citep{Henaff2017,Santoro2017} or visual inputs
\citep{Santoro2017}.
Here, we take a different approach, and consider a logic-based formal reasoning problem as a
starting point instead.
The choice of the particular problem turns out to be a critical one, though.
While logic-based formal reasoning defines a certain conceptual frame, the specific nature of and
extent to which inferences may be drawn are highly dependent on the concrete formalism employed.
Therefore, it is generally sensible to choose an approach that presents the right balance between
expressiveness on the one hand and complexity on the other hand.
This criterion as well as its vast importance in practice made us look into the problem of
ontology reasoning with OWL 2 RL.
Ontology reasoning refers to a common scenario where the inference rules to be used for reasoning,
called the ontology in this context, are specified along with the factual information that we
seek to reason about.
Figure~\ref{fig:okb-example} provides an example of this setting.
The rationale behind this separation of ontology and facts is that it allows for adopting the same
set of rules for reasoning about different, independent data.
What makes this task particularly interesting, though, is that the application of generic rules to
concrete problem settings happens to describe a reoccurring pattern in all of our lives, which is
why human beings are in general very good at handling problems of this kind.
Furthermore, ontology reasoning is an incredibly pliant tool, which allows for modeling a plethora
of different scenarios, and as such meets our desire for a system that is applicable to a wide range
of applications.

\begin{figure}[t]
    \centering
    {
        \setlength{\tabcolsep}{2pt}
        \footnotesize
        
        \begin{tabular}{
            >{\bfseries}       l <{\hspace{0.25cm}}
            >{\ttfamily}       l
                               l
            >{\ttfamily}       l
            >{\hspace{0.25cm}} l
        }
            
            Ontology:
                &human(X)  & $\gets$ & holds(X,\_)
                &Only human beings can hold things.                                              \\
                
                &object(Y) & $\gets$ & holds(\_,Y)
                &Only objects can be held.                                                 \\
                
                &$\bot$    & $\gets$ & human(X)\,$\wedge$\,object(X)
                &Objects are not human beings and vice versa.                                    \\
                
                &isAt(Y,Z) & $\gets$ & holds(X,Y)\,$\wedge$\,isAt(X,Z)
                &Objects are at the same location as the one                               \\
                &&&&holding them.                                                          \\
                
                &$\bot$    & $\gets$ & isAt(X,Y)\,$\wedge$\,isAt(X,Z)\,$\wedge$\,Y$\neq$Z
                &Nobody/nothing can be at two locations at                                 \\
                &&&&the same time.                                                         \\
            \\
            Facts:
                &\multicolumn{3}{l}{\ttfamily holds(mary,apple)}
                &Mary holds the apple.                                                     \\
                
                &\multicolumn{3}{l}{\ttfamily isAt(mary,kitchen)}
                &Mary is in the kitchen.                                                   \\
            \\
            Queries:
                &\multicolumn{3}{l}{\ttfamily ?human(apple)}
                &Is the apple a human being?                                               \\
                &&&&(Evaluates to\ \textbf{false}.)                                        \\
                
                &\multicolumn{3}{l}{\ttfamily ?isAt(apple,kitchen)}
                &Is the apple in the kitchen?                                              \\
                &&&&(Evaluates to\ \textbf{true}.)                                         \\
                
                &\multicolumn{3}{l}{\ttfamily ?isAt(mary,bedroom)}
                &Is Mary in the bedroom?                                                   \\
                &&&&(Evaluates to\ \textbf{false}.)                                        \\
            \\
        \end{tabular}
    }
    
    \caption{%
        This figure provides a simple example of an ontology, which was inspired by the well-known
        bAbI tasks \citep{Weston2015a}.
        An ontology is a collection of generic rules, which are combined with a set of facts.
        Here, the ontology describes a few rules for reasoning over human beings, objects, and their
        locations.
        Combined with the stated facts, it allows for answering queries like
        \MyQuote{Is the apple a human being?} or \MyQuote{Is Mary in the bedroom?}.
    }
    \label{fig:okb-example}
\end{figure}

One may ask why we would like to set about this problem by means of machine learning in the first
place.
For one thing, such combinations of deep learning technologies with symbolic methods for logical
reasoning are commonly regarded as a prerequisite for further substantial progress in AI.
In particular, they are expected to allow for leveraging symbolic background knowledge in learning
deep neural networks (and so for knowledge transfer and for learning from smaller amounts of
data) as well as explainable symbolic inference in computing predictions for improved explainability of the learned neural systems.
Furthermore, most of the KRR formalisms that are used for reasoning today are rooted in symbolic
logic, and thus, as mentioned above, employ mathematical proof theory to answer queries about a
given problem.
However, while this, in theory, allows for answering any kind of (decidable) question accurately,
most of these approaches suffer from a number of issues in practice, like difficulties with handling
incomplete, conflicting, or uncertain data, to name just a few.
In contrast to this, machine learning models are often highly scalable, more resistant to
disturbances in the data, and capable of providing predictions even if the formal effort fails.
There is one salient aspect, though, that KRR methods benefit from compared to learning-based
approaches: every conclusion derived via formal inference is correct with certainty,
and, under optimal circumstances, formal reasoning identifies all inferences that can be drawn
altogether.
These characteristics do, in general, not apply to methods of machine learning.
However, \MyQuote{optimal} is the operative word right here, as these advantages can often not come
into play due to the obstacles mentioned above.
By employing state-of-the-art techniques of deep learning, we aim to manage the balancing act
between approximating the highly desirable (theoretical) properties of the formal approach, on the
one hand, and utilizing the robustness of machine learning, on the other.

The main contributions of this paper are summarized as follows: 
\begin{itemize}
    \item We present a novel deep neural architecture for a model that is able to effectively
          perform logical reasoning in the form of basic ontology reasoning.
    \item We present, and make freely available, several very large, diverse, and challenging
          datasets for learning and benchmarking machine learning approaches to basic ontology
          reasoning.
    \item We present extensive experimental evaluations on the above benchmarks, which show that our
          model is able to learn to perform highly accurate ontology reasoning.
\end{itemize}

\raggedbottom  

The rest of this article begins with a description of the reasoning problem that served as the
starting point of our research.
After this, we introduce the RRN model, explain how it has been evaluated in several experiments,
and present the outcomes of our experimental evaluation.
Finally, we review relevant related work, and conclude with a discussion.


\section{Problem Description}



The major part of the knowledge bases that are used for ontology reasoning today formalizes
information in terms of individuals, classes, and binary relations, any of which may thus be
considered as a directed knowledge graph, where individuals correspond to vertices, relations to
labeled directed edges, and classes to binary vertex labels.\footnote{%
    In the context of relational learning, knowledge graphs are typically simplified by viewing
    classes as individuals as well and memberships as ordinary relations.
    For our purposes, however, a clear distinction between classes and relations is important, which
    is why we adopt this slightly different view here.
}
In KRR terminology, a setting like this may be described as a basic ontological knowledge base with
a function-free signature that contains unary and binary predicates only.
The facts that define such a knowledge graph are usually stated in terms of triples of the form
$\langle \mathit{subject}, \mathit{predicate}, \mathit{object} \rangle$,
and specify either a relation between two individuals, \subject\ and \object, or an
individual's membership of a class, in which case \subject\ refers to an individual, \predicate\ to
a special membership relation, and \object\ to a class.
Notice that we assume a fixed vocabulary, which means that all of the considered classes and
relations are known beforehand, and are thus regarded as part of the ontology.
The rules of the ontology are usually specified in terms of a knowledge representation language,
such as a logic program or OWL.


We now provide a formal description of the reasoning problem that we consider in this article, and
thus put our work in a formal context.
Throughout this paper, we consider ontology reasoning that corresponds to an (extension of~a) subset
of Datalog where relations have an arity of at most two, but are not partitioned into database and
derived relations, and where rules with falsity $\bot$ in the head, called
\emph{negative constraints}, are allowed.
To that end, we assume an infinite set of \emph{constants}~$\Constants$, an infinite set of
\emph{variables}~$\Variables$, and a \emph{relational schema} $\Schema$, which is a finite set of
names for unary and binary relations.
A \emph{term}~$t$ is a constant or a variable, and an \emph{atomic formula} (or \emph{atom}) has the 
form~of either $p(t)$ or $p(t_1,t_2)$, where $p$ is a relation name, and $t$, $t_1$, and $t_2$ are
terms.
A~\emph{rule}~$r$ has the form 
\begin{equation}\label{eq:rule}
    \begin{array}{l}
        \BodyVar{1} \wedge \cdots \wedge \BodyVar{n}
        \rightarrow
        \HeadVar
        \text{,}
    \end{array}
\end{equation}
where $\HeadVar, \BodyVar{1},\ldots, \BodyVar{n}$ are atoms and $n\,{\ge}\, 0$.
Such a rule is \emph{safe}, if each of the variables in $\HeadVar$ also occurs in at least
one~$\BodyVar{i}$ (for $1 \leq i \leq n$).
A~\emph{negative constraint}~$\Constraint$ has the form 
\begin{equation}\label{eq:rule2}
    \begin{array}{l}
        \BodyVar{1} \wedge \cdots \wedge \BodyVar{n}
        \rightarrow
        \bot
        \text{,}
    \end{array}
\end{equation}
where $\BodyVar{1},\ldots, \BodyVar{n}$ are atoms and $n\,{\ge}\, 0$.
A \emph{program} $\Ontology$ is a finite set of safe rules of the form~(\ref{eq:rule}) and negative
constraints of the form~(\ref{eq:rule2}) with $n\ge 1$.
A safe rule with $n = 0$ is a \emph{fact}, and a \emph{database} is a finite set of facts.
A \emph{literal}~$\ell$ is either a fact~$\alpha$ or a negated fact~$\neg \alpha$. 

A (Herbrand) \emph{interpretation} $\Interpretation$  is a (possibly infinite) set of facts.
An interpretation $\Interpretation$ \emph{satisfies} a variable-free rule $r$ of the
form~(\ref{eq:rule}), if $\{\BodyVar{1},\ldots,\BodyVar{n}\} \subseteq \Interpretation$ implies
$\alpha \in I$.
An interpretation $\Interpretation$ \emph{satisfies} a variable-free negative constraint
$\Constraint$ of the form~(\ref{eq:rule2}), if
$\{\BodyVar{1},\ldots,\BodyVar{n}\} \not \subseteq \Interpretation$.
Furthermore,  $\Interpretation$ \emph{satisfies} a set $R$ of rules and negative constraints, if it
satisfies all their variable-free instances.
We say that $R$ is \emph{satisfiable}, if it has a satisfying interpretation.
Finally, a fact~$\zeta$ (resp., negated fact $\neg\zeta$) is \emph{logically entailed} by~$R$,
denoted  $R \models \zeta$ (resp., $R \models \neg \zeta$), if it is part (resp., not part) of every
interpretation that satisfies $R$.
For satisfiable sets $R$ of rules and negative constraints, since each such $R$ has a unique least
satisfying interpretation, denoted~$M_R$, and  this is equivalent to $\zeta \in M_R$ (resp.,
$\{\zeta\} = \{\BodyVar{1},\ldots,\BodyVar{n}\}\setminus M_R$ for some variable-free instance of the
form~(\ref{eq:rule2}) of a negative constraint in $R$).
In the sequel, we implicitly assume that all considered sets $R$ of rules and negative constraints
are satisfiable.

Further negated facts can be derived, if we additionally make the
\emph{closed-world assumption}~(CWA) or the \emph{local CWA}~(LCWA).
Formally, a negated fact~$\neg\alpha$ is \emph{logically entailed by $R$ under the CWA}, if and only
if $\alpha$ is not entailed by the same, that is, $R \models \neg\alpha$ if and only if
$R \not\models \alpha$.
\emph{Logical entailment under the LCWA} is a restricted subset of logical entailment under the CWA
and superset of logical entailment that additionally allows for deriving negated atoms about binary
predicates that are similar (\MyQuote{local}) to entailed positive facts in the following sense: if
$R \not\models p(u_1,u_2)$ and $R \not\models \neg p(u_1,u_2)$, then, under the LCWA,
$R \models \neg p(u_1,u_2)$ if and only if either $t_1 = u_1$ or $t_2 = u_2$ for some $p(t_1,t_2)$
with $R \models p(t_1,t_2)$.
We denote by~\MyQuote{$\Entails$}  one of these three logical entailment relations.

We are now ready to define the problem that this article aims to solve, the
\emph{logical entailment problem} (of facts from databases and programs):
given a database~$D$, a program~$\Ontology$, and a literal $\ell$, decide whether
$\Ontology\cup D\,\Entails \, \ell$.
In this work, we consider the most common case where $D$ is variable and of size $k$, $\Ontology$ is
fixed, and $\ell$ is variable as well.
The goal is to generate a neural network~$N[{\Ontology,k}]$ with binary output that, given an
arbitrary database $D$ of size at most $k$ as well as an arbitrary literal~$\ell$, decides the
logical entailment problem, that is, $N[{\Ontology,k}](D,\ell)\,{=}\,1$  if and only if
$\Ontology\cup D\,\Entails \, \ell$.


\section{The Recursive Reasoning Network (RRN)}

To tackle the problem specified in the previous section, we introduce a novel model archi\-tecture,
the \emph{recursive reasoning network} (RRN).
To that end, this section starts with a high-level outline of the RRN model, and then provides a
detailed description of the same.


\subsection{Intuition}

With the introduction of RRNs, we replace formal ontology reasoning with computing a learned deep
neural network to remedy the issues outlined above.
Thereby, following the spirit of the considered problem, every RRN is trained relative to a
particular ontology, and is thus, like the formal counterpart, independent of the specific facts
that it is provided with.
To that end, as described in detail below, the vocabulary of classes and relations used by the
ontology determines the recursive layers that are available in an RRN, and hence the structure of
the same.
In contrast to this, however, the rules that are used for reasoning are not provided to the model
directly, but have to be learned from the training data.
When a trained model is applied to a particular set of facts, then, on the face of it, it operates
in two stages (each of which is explained subsequently): first, it generates vector representations,
so-called embeddings, for all individuals that appear in the considered data, and second, it
computes predictions for queries solely based on these generated vectors.

RRNs are based on the idea that we can encode all the information that we have about an
individual, both specified and inferable, in its embedding.
A similar idea is employed, for example, in the context of natural language processing, where real
vectors are used to represent the meaning of text \citep{Mikolov2013}.
Given a set of facts, specified as triples, we start by randomly generating initial embeddings for
all the individuals that appear in any of them.
After this, the model iterates over all the triples, and, for each of them, updates the
embeddings of the individuals involved.
Any such update is supposed to store the considered triple in the individuals' embeddings for one
thing, but also to encode possible inferences based on the same.
So, intuitively, a single update step conducts local reasoning based on what is encoded in the
embeddings already as well as on the new information that was gained through the provided fact.
An obvious necessity implied by this local reasoning scheme is that the model, in general, has to
sift through all  the data multiple times.
This is essential in order to allow for multi-step, also called multi-hop, reasoning, which is based
on several triples at the same time, since information might need to transpire from one individual's
embedding to another's.
The actual number of iterations required depends on the respective dataset, though.
Figure \ref{fig:embedding-generation} summarizes this process.

From a technical perspective, the outlined procedure for generating embeddings corresponds to 
computing a recursive neural network \citep{Pollack1990} that receives the randomly generated
initial embeddings as input and provides the final embeddings as output.
The structure of the network depends on the set of facts being processed, and its recursive layers
compute the update operations described above, which is why we refer to them as update layers.

\ModelFigure

Once the desired embeddings are generated, they can be used to answer atomic queries about the data that they are computed from.
To that end, the model provides various multi-layer perceptrons (MLPs) for computing predictions
about relations between two individuals and class memberships of a single individual.
Notice that the only inputs provided to these MLPs are the embeddings of the individuals that are
involved in a particular query, which is why the model has to ensure that all the information
that is needed for answering such is effectively encoded during the first step.
Thus, the second step is just needed to uncover the knowledge that is encoded in individual
embeddings, while the actual reasoning happens before.

A notable characteristic of the RRN is that it performs deductive inference over entire knowledge
bases, and, like its symbolic stencil, aims to encode \emph{all} possible inferences in the created
individual embeddings, rather than answering just a single query.
Because of this, the model is able to unravel complex relationships that are hard to detect if we
try to evaluate the inferability of an isolated triple of interest only.
Furthermore, the fact that the RRN conducts logical inference over all classes and relations
simultaneously allows for leveraging interactions between any of them, and thus further adds to
improving the model's predictive performance.

Another noteworthy aspect is that, as we will see below, an RRN does \emph{not} treat triples as
text.
Instead, the individuals that appear in a triple are mapped to their embeddings before the same is
provided to any layers of the used model.
However, this means that RRNs are agnostic to the individual names that are used in a database,
which makes perfect sense, as it is only the structure of a knowledge base that determines possible
inferences.

    
\subsection{Terminology}

In the sequel, we always assume that we are facing a dataset that is given as an ontological
knowledge base~$\OKB = \langle \Ontology,\Facts \rangle$, consisting of  a program $\Ontology$
(also called the \emph{ontology}) and a database $\Facts$, as defined above.
We do not impose any restrictions on the formalism that has been used to specify $\OKB$, but we do
require that it makes use of a fixed vocabulary that contains unary and binary predicates only,
which we refer to as classes and relations, respectively, and that the individuals considered have
been fixed as well.
The sets of all classes and individuals of $\OKB$ are denoted as $\Classes ( \OKB )$ and
$\Individuals ( \OKB )$, respectively.
Notice that these assumptions allow for viewing $\Facts$ as a set of triples, since we may represent
any relation $\RelationSymbol ( i, j )$ as $\Triple{i}{\RelationSymbol}{j}$ and any class membership 
$\ClassSymbol ( i )$ as $\Triple{i}{\Member}{C}$, where $\Member$ is a special symbol that does not
appear in the vocabulary.
Furthermore, we can represent negated facts as triples with a different relation symbol, e.g.,
$\neg \RelationSymbol ( i, j )$ and $\neg \ClassSymbol ( i )$ as
$\Triple{i}{\neg \RelationSymbol}{j}$ and $\Triple{i}{\neg \Member}{C}$, respectively.

For every knowledge base $\OKB$, we define an indicator function
$$
    \Incidence{\OKB} : \Individuals ( \OKB ) \to \{ -1, 0, 1 \}^{| \Classes ( \OKB ) |}
$$
that maps the individuals that appear in $\OKB$ to 3-valued incidence vectors.
Any such vector summarizes all the information about the respective individual's class
memberships that are given \emph{explicitly} as part of the facts~$\Facts$.
So, if the considered classes are specified via the predicates
$\Class{1} , \Class{2} , \dotsc , \Class{\NumClasses}$
and $i \in \Individuals ( \OKB )$, then
$\Incidence{\OKB} ( {i} )$ yields an $m$-dimensional vector, and 
for $\ell \in\{ 1 , \dotsc , \NumClasses\}$,
$$
    \big[ \Incidence{\OKB} ( {i} ) \big]_\ell =
    \left\{
        \begin{array}{r l}
            1     &\text{if $\Triple{i}{\Member}{\Class{\ell}} \in \Facts$,}             \\
            -1    &\text{if $\Triple{i}{\neg \Member}{\Class{\ell}} \in \Facts$, and}    \\
            0     &\text{otherwise.}                                                     \\
        \end{array}
    \right.
$$

    
\subsection{Formal Definition of the Model}

The purpose of the RRN model is to solve the logical entailment problem
defined above, and to that end, we train an RRN to reason relative to a fixed ontology.
More formally, suppose that $\RRN{\Ontology}$ is a model that has been trained to reason with
the ontology $\Ontology$, and~$\Facts$, like before, denotes a finite set of facts.
Furthermore, let the triple $T = \Triple{s}{P}{o}$ be an arbitrary query whose entailment is to be
checked.
Then, the network $\RRN{\Ontology}$ defines a function such~that
\begin{equation}\label{eq:rrn}
    \RRN{\Ontology} ( \Facts , T ) =
    \Prob{\text{$T$ is true} \mid \langle \Ontology , \Facts \rangle}
    \text{.}
\end{equation}
There are a few important aspects to notice about Equation \ref{eq:rrn}.
First and foremost, $\Facts$ may be any set of facts that use the same vocabulary as $\Ontology$,
and does \emph{not} have to be the one that has been used to train $\RRN{\Ontology}$.
Similarly, $T$ may be an arbitrary triple, once again, sharing the same vocabulary, and does
\emph{not} have to appear in the training data---neither as fact nor as inference.
This means that $\RRN{\Ontology}$ actually performs ontology reasoning, and does not just
\MyQuote{memorize} the ontological knowledge base(s) that it has been trained on.
The probability on the right-hand side of Equation \ref{eq:rrn} is used to express the model's
belief in whether the queried triple is true based on the ontological knowledge base that is
formed implicitly by the ontology that the model was trained on together with the provided set of
facts, that is, $\langle \Ontology , \Facts \rangle$.
Interpreting the model's output as a probability allows for training it via a cross-entropy error
term, but during evaluation, we predict queries to be true as soon as the model provides a
probability of at least 50\%.
Note that we deliberately chose the wording  \MyQuote{$T$ is true}, rather than
$\Ontology \cup \Facts \, \Entails \, T$, as an RRN provides a prediction even if $T$
is neither provably entailed nor refutable based on $\langle \Ontology , \Facts \rangle$.

We illustrate these ideas with a simple example.
Suppose that we require a model for reasoning with respect to some ontology $\Ontology$.
Then, as a first step, we train an RRN for reasoning with this particular ontology, denoted
$\RRN{\Ontology}$.
To train this model, we can either use an existing database at hand or simply generate (consistent)
sets of facts that make use of the same vocabulary, that is, the same set of predicate symbols, as
training samples.
After this, $\RRN{\Ontology}$ can be used to perform ontology reasoning (relative to~$\Ontology$)
over arbitrary databases that use the same vocabulary as~$\Ontology$.
For instance, suppose that we need to check whether
$\Ontology \cup D_1 \, \Entails \, T$
for some database $D_1$ and triple $T$.
In this case, we  employ $\RRN{\Ontology}$ to first generate an embedding matrix~$\Mat{E}_{D_1}$ for the
individuals in $D_1$, and second compute a probability for whether the considered entailment holds
from $\Mat{E}_{D_1}$.
Taken together, these two steps constitute the function specified in Equation~\ref{eq:rrn}.
We emphasize again that the actual reasoning step happens as part
of the embedding generation, and entailment probabilities are computed solely based on generated
embedding matrices.
Notice further that $\RRN{\Ontology}$ is not tied to $D_1$ in any way, which means that we can use
the same model to perform ontology reasoning over another (different) database $D_2$.
Finally, note that, in practice, we do not generate new embeddings whenever we check an entailment
based on one and the same set of facts, but instead, create and reuse just a single embedding matrix
for each database.

    
\subsection{Individual Embeddings}

We represent individuals as unit vectors, with respect to the Euclidean norm, of the 
space~$\R^\EmbeddingSize$, which means that all embeddings live in the $\EmbeddingSize$-dimensional unit
hypersphere.
To that end, $d$ is a hyperparameter that, in general, has to be chosen based on the
expressiveness of the ontology, the size of the used vocabulary, and the total number of individuals
considered.
In this work, we denote the embedding of an individual $i$ as $\Embedding{i}$.

Initially, individual embeddings are generated randomly following some probability distribution,
which was chosen to be a uniform distribution in our experiments, and normalized subsequently,
that is,
$$
    \Candidate{i_1}, \Candidate{i_2}, \dots \overset{\text{iid}}{\sim}
    \UniformDist{-\Vect{1}_\EmbeddingSize}{\Vect{1}_\EmbeddingSize}
    \qquad
    \text{and}
$$
$$
    \Embedding{i_\ell} = \frac{\Candidate{i_\ell}}{\Norm{\Candidate{i_\ell}}}
    \qquad
    (1 \leq \ell \leq | \Individuals ( \OKB ) |)
    \text{,}
$$
for $\Individuals ( \OKB ) = \{ i_1, i_2, \dots \}$.

    
\subsection{Update Layers}

At the heart of the RRN model, there are two kinds of update operations, one based on relations and
one based on class memberships.
Both of these are very similar in nature, and serve the same essential purpose: given a fact
together with  the embeddings of the individuals involved, update these embeddings to incorporate
the knowledge that was gained through the provided fact (under the used ontology~$\Ontology$ as well
as the information that is encoded in the embeddings already).

We start with the update operation for relations, which means that we aim to update the embeddings
$\Embedding{s}$ and $\Embedding{o}$ based on some triple $\Triple{s}{P}{o}$ with $P$ being a
relation type, that is, neither $\Member$ nor $\neg \Member$.
Notice, however, that $P$ might as well be a negated relation type $\neg \RelationSymbol$.
To define a proper update operation, one has to mind the fact that relations are, in general, not
symmetric, and hence it makes a difference whether we update a triple's subject or object.
Therefore, we define two (equal) recursive update layers for every (possibly negated) relation type
$P$ that appears in $\OKB$, one for updating the subject's embedding, which we denote as
$\text{\it Update}^{\triangleleft P}$, and one for updating the object's, denoted 
$\text{\it Update}^{P \triangleright}$.
The following equations describe
$\text{\it Update}^{\triangleleft P}$---$\text{\it Update}^{P \triangleright}$ is defined
analogously:
\begin{equation}\label{eq:gate}
	\GateSourceUpdate{s}{P}{o} =
    \sigma \Big(
        \GateSourceUpdateWeightsSource{P} \Embedding{s} +
        \GateSourceUpdateWeightsTarget{P} \Embedding{o}
    \Big)
    \text{,}
\end{equation}
\begin{equation}\label{eq:update}
    \Candidate{s}^{( 1 )} =
    \ReLU \Big(
        \UpdateWeightsSource{\triangleleft P} \Embedding{s} + 
        \UpdateWeightsTarget{\triangleleft P} \Embedding{o} + 
        \Embedding{s} {\Embedding{o}}^T \UpdateWeightsInteraction{\triangleleft P}
    \Big)
    \text{,}
\end{equation}
$$
    \Candidate{s}^{( 2 )} =
    \Embedding{s} + \Candidate{s}^{( 1 )} \circ \GateSourceUpdate{s}{P}{o}
    \text{,}
$$
$$
    \Embedding{s} =
    \UpdateSource{s}{P}{o} =
    \frac{\Candidate{s}^{( 2 )}}{\Norm{\Candidate{s}^{( 2 )}}}
    \text{.}
$$
As usual, $\sigma ( x )$ denotes the logistic function $1 / (1 + \exp(-x))$ and $\ReLU ( x )$ the
ramp function $\max \{ 0, x \}$, both of which are applied elementwise.
$\circ$ denotes the elementwise vector product, and
$
    \GateSourceUpdateWeightsSource{P},
    \GateSourceUpdateWeightsTarget{P},
    \UpdateWeightsSource{\triangleleft P},
    \UpdateWeightsTarget{\triangleleft P}
    \in \R^{d \times d}
$ as well as
$\UpdateWeightsInteraction{\triangleleft P} \in \R^d$
are parameters of the model.
As can be seen from these equations, the recursive layers of the RRN model define a gated network
architecture.
To that end, Equation \ref{eq:update} describes the calculation of a \MyQuote{candidate} update
step, and Equation \ref{eq:gate} specifies a gate that controls how much of this is actually
applied.
Based on these one-sided updates, we can now define the following simultaneous update operation:
$$
    \langle \Embedding{s}, \Embedding{o} \rangle =
    \FullUpdate{s}{P}{o} =
    \big\langle \UpdateSource{s}{P}{o}, \UpdateTarget{s}{P}{o} \big\rangle
    \text{.}
$$

Next, we consider updates based on class memberships.
In this case, there are no interactions between individuals, which is why updates of this kind may
be batched.
Also, we do not have to pay special attention to the \MyQuote{direction} of an update, neither to
whether a membership assertion is positive or negative.
Instead, we simply define a single recursive update layer, which expects an individual $i$ as input,
as follows:
$$
    \Vect{g} ( i ) =
    \sigma \Big(
        \Mat{V} \cdot [ \Embedding{i} : \Incidence{\OKB} ( {i} ) ]
    \Big)
    \text{,}
$$
$$
    \Candidate{i}^{( 1 )} =
    \ReLU \big(
        \Mat{W} \cdot [ \Embedding{i} : \Incidence{\OKB} ( {i} ) ]
    \big)
    \text{,}
$$
$$
    \Candidate{i}^{( 2 )} =
    \Embedding{i} + \Candidate{i}^{( 1 )} \circ \Vect{g} ( i )
    \text{,}
$$
$$
    \Embedding{i} =
    \text{\it Update} ( i ) =
    \frac{\Candidate{i}^{( 2 )}}{\Norm{\Candidate{i}^{( 2 )}}}
    \text{.}
$$
This layer updates an individual's embedding based on \emph{all} the information that has been
provided about its class memberships.
Notice that, in these equations, the colon denotes the concatenation of two vectors.
Like before, the layer makes use of a gated architecture, and
$\Mat{V}, \Mat{W} \in \R^{d \times ( d + | \Classes ( \OKB ) | )}$
are parameters of the model.

\begin{algorithm}[t]
    \footnotesize
    \SetAlgoLined
    
    \KwIn{%
        an ontological knowledge base
        $\OKB = \langle \Ontology , \Facts \rangle$ with
        $\Individuals ( \OKB ) = \{ i_1 , i_2 , \dotsc , i_M \}$, a~number of update iterations $N$, and (optionally) a matrix of initial embeddings $\Mat{E}$. 
    }
    
    \KwOut{the generated embeddings $\Mat{E}$.\vspace*{1ex}}

    \If{no embedding matrix $\Mat{E}$ was provided}{%
        Randomly initialize
        $\Mat{E} = [ \Embedding{i_1}, \Embedding{i_2}, \dotsc , \Embedding{i_M} ]^T$\;
    }
    \For{$\text{\it iter} = 1 , \dotsc , N$}{%
        \ForEach{$i \in \Individuals ( \OKB )$}{%
            $\Embedding{i} = \text{\it Update} ( i )$\;
        }
        \ForEach{$\Triple{s}{P}{o} \in F$ with $P$ being a (possibly negated) relation type}{%
            $\Embedding{s}, \Embedding{o} = \FullUpdate{s}{P}{o}$\;
        }
    }
    \Return{$\Mat{E}$}.
    
    \caption{Generating individual embeddings.}
    \label{alg:embedding-generation}
\end{algorithm}

Algorithm \ref{alg:embedding-generation} summarizes how these layers are used to generate embeddings
for all individuals in the considered knowledge base.


\subsection{Prediction}

For computing predictions based on previously generated individual embeddings, we make use of
several MLPs, a single one for all classes and one for each relation $R$ in the vocabulary,
denoted $\MlpClasses$ and $\MlpRelations{\RelationSymbol}$, respectively.
Notice that, in this case, $R$ is indeed a relation type, rather than its negation.
$\MlpClasses$ expects a single individual embedding as input, and computes probabilities for the
individual's memberships with respect to all classes.
If we assume that $\OKB$ contains the classes $\ClassSymbol_1 , \dotsc , \ClassSymbol_\NumClasses$,
this means that $\MlpClasses$ computes a function
$$
    \MlpClasses : \R^d \to [ 0 , 1 ]^\NumClasses
$$
and
$$
    \big[ \MlpClasses ( \Embedding{i} ) \big]_\ell =
    \Prob{\Triple{i}{\Member}{\ClassSymbol_\ell} \mid \OKB}
    \text{.}
$$
Similarly, $\MlpRelations{\RelationSymbol}$ requires two individual embeddings as input, and
provides a probability for the according relation to exist between the respective individuals.
Assuming that $\OKB$ consists of the relations
$\RelationSymbol_1 , \dotsc , \RelationSymbol_\NumRelations$, this means that
$$
    \MlpRelations{\RelationSymbol_\ell} : \R^{2 d} \to [ 0 , 1 ]
$$
and
$$
    \MlpRelations{\RelationSymbol_\ell} ( \Embedding{i}, \Embedding{j} ) =
    \Prob{\Triple{i}{\RelationSymbol_\ell}{j} \mid \OKB}
    \text{.}
$$
Notice that, at the same time,
$$
    \Prob{\Triple{i}{\neg \RelationSymbol_\ell}{j} \mid \OKB} =
    1 - \Prob{\Triple{i}{\RelationSymbol_\ell}{j} \mid \OKB} =
    1 -         \MlpRelations{\RelationSymbol_\ell} ( \Embedding{i}, \Embedding{j} )
    \text{.}
$$


\subsection{Training}
\label{sec:training}

The training of an RRN is straightforward, and a high-level description is provided in
Algorithm~\ref{alg:rrn-training}.
An important detail to notice is that all training samples are knowledge bases that share the same
ontology, namely, the one that the RRN is being trained for, but usually differ with respect to the
facts that they contain.
In contrast to this, many approaches to learning-based reasoning and knowledge-base completion make
use of one large knowledge graph for training a model only, which is usually also the one that has
to be completed.
As stressed above already, this is a major advantage of the RRN, since it allows for training a
model with respect to an ontology, but independently of the knowledge base(s) that it is supposed to
be applied to later on.
To that end, one may simply \emph{generate} sample knowledge bases relative to the ontology of
interest, and train the RRN on this synthetic dataset.
Once a model has been trained, it can be applied directly to any previously unseen knowledge base
without the need for being retrained.
Another important advantage of the possibility to train on generated data is that it allows for the
model to encounter predicates during training that appear as inferences, but not as facts in the
knowledge base that it is eventually applied to.
As opposed to this, models that are trained on the same knowledge base that has to be completed have
no means to learn about such predicates, and thus cannot identify according inferences.

Intuitively, this approach might not seem sound, as it strongly disagrees with the usual way of
training a model, and one may challenge whether an RRN is trained on the right data distribution
this way.
We argue that this is indeed the case, though.
Training a model to perform ontology reasoning requires the same to learn how to make use of the
rules comprised by the considered ontology.
Furthermore, an RRN is not provided these rules directly, but has to learn them from the training
data as well.
Notice, however, that the application of any such rule is exactly the same for a synthetically
generated knowledge base and another one encountered as part of a real-world problem, assuming that
both rely on the same ontology, and thus there is no reason to favor real-world data in this
context.
In fact, synthetic datasets offer the possibility for us to control that all rules in an ontology
are used to, more or less, the same extent, while real-world knowledge bases tend to be skewed
towards a part of the rules only.
Therefore, we have to make sure that all of the ontology is faithfully represented by a real-world
dataset, while a sufficiently large synthetic one is clearly capable of accomplishing this.

Besides that, the training procedure for an RRN closely resembles the common pattern for training
supervised models.
This means that we sample a knowledge base from the training data, generate individual embeddings
for the same, and compute predictions for all facts and inferences that can be drawn from the
considered knowledge base.
Notice that every fact in a knowledge base is trivially entailed by the same, and hence, strictly
speaking, part of the inferences that the model learns to predict.
For evaluating our approach, a clear distinction of facts and inferences is useful, though, which is
why we use the terms \MyQuote{inference} and \MyQuote{inferable} in the sense of inferable but not
specified as fact.
Finally, we compute a cross-entropy loss for the computed predictions, and adjust the model via
\ac{SGD}.
After every training epoch, the model is evaluated on a hold-out set, and the training procedure is
stopped as soon as the error that was computed on the same has converged.

\begin{algorithm}[t]
    \footnotesize
    \SetAlgoLined
    
    \KwIn{%
        a sequence of training samples
        $T = \langle \OKB_1 , \OKB_2 , \dotsc , \OKB_M \rangle$, where
        $\OKB_\ell = \langle \Ontology , \Facts_\ell \rangle$, 
        and a~number of update iterations $N$.\vspace*{1ex}
    }

    \While{evaluation error has not converged}{%
        Randomly shuffle $T$\;
        \For{$\OKB_\ell \in T$}{%
            Generate embedding matrix $\Mat{E}$ for $\OKB_\ell$\;
            Compute cross-entropy loss for predicting triples in $\OKB_\ell$ from $\Mat{E}$
            (both facts and inferences)\;
            Update model parameters to minimize the loss\;
        }
    }
    
    \caption{RRN training.}
    \label{alg:rrn-training}
\end{algorithm}

A salient detail of the outlined training procedure is that we consider just a single training
sample at a time, and thus seem to use an online version of \ac{SGD}.
However, as samples are knowledge bases, they usually induce a multitude of triples that have to be
predicted in each training iteration, and hence effectively correspond with minibatches of varying
sizes.


\subsection{Computational Aspects}
\label{sec:complexity}

Now that we have defined all parts of the RRN model, we need to ask the question of whether the
presented approach is sensible from a practical point of view.
For most  problems that we encounter in practice, we can consider the ontology $\Ontology$
to be fixed, while the database of known facts $\Facts$ may be subject to frequent change.
Therefore, while we can safely disregard the cost for training a model $\RRN{\Ontology}$, the
complexity of computing the same in order to reason over $\Facts$ (relative to $\Ontology$) turns
out to be critical.

We start with investigating the cost of generating individual embeddings.
As stated in Algorithm \ref{alg:embedding-generation}, each triple in $\Facts$ triggers (at most)
one update operation per update iteration.\footnote{%
    In Algorithm \ref{alg:embedding-generation}, updates based on triples that specify information
    with respect to class memberships (lines 5 to 7) are grouped by individual and batched.
    In practice, we would perform such a batch update only if there is at least one fact concerning
    the respective individual.
    This means that, in the worst case, we have to perform one update for each fact about a class
    membership per update iteration.
}
Since every update can be computed in constant time and the number of update iterations is fixed,
this means that the (time) cost for generating an embedding matrix $\Mat{E}_\Facts$ is (at most)
linear in the size of $\Facts$, that is, $\BigO{|\Facts|}$.
This can be further improved by batching update operations for triples that describe facts
about relations, and parallelizing the computation of any such batch of updates.
One way to find appropriate batches is to compute maximal matchings in the considered knowledge
graph restricted to a single relation type, which can be realized by means of fast greedy
algorithms.
Notice further that, as pointed out above already, we usually do not compute new embeddings every
time we need to check an entailment, but only if the database $\Facts$ changed.

The second operation involved in using $\RRN{\Ontology}$ is computing predictions from
$\Mat{E}_\Facts$.
However, this requires constant time for each query, and once again, multiple queries may be batched
in order to further reduce the time of computation.


\section{Evaluation}

\begin{table}[t]
    \centering
    \footnotesize
    
    \setlength{\tabcolsep}{2.9pt}

    \begin{tabular}{| l | c c | c | c c |}
    
        \hline
        
        \multicolumn{1}{| c |}{\textbf{dataset}}            &
            \multicolumn{2}{c |}{\textbf{\# sample KBs}}    &
            \textbf{avg. ind.}                              &
            \multicolumn{2}{c |}{\textbf{vocabulary size}}  \\
        
        &
            \textbf{train}              &
            \textbf{test}               &
            \textbf{per sample}         &
            \textbf{\# class types}     &
            \textbf{\# relation types}  \\ \rowcolor{\RowColor}
        
        \hline
        
        family trees    &  5,000  &  500  &  23   &  2    &  29   \\
        countries (S1)  &  5,000  &  20   &  240  &  3    &  2    \\ \rowcolor{\RowColor}
        countries (S2)  &  5,000  &  20   &  240  &  3    &  2    \\
        countries (S3)  &  5,000  &  20   &  240  &  3    &  2    \\ \rowcolor{\RowColor}
        DBpedia         &  5,000  &  500  &  200  &  101  &  518  \\
        Claros          &  5,000  &  500  &  200  &  33   &  77   \\ \rowcolor{\RowColor}
        UMLS-reasoning  &  5,000  &  500  &  60   &  127  &  53   \\
        
        \hline
    
    \end{tabular}
    
    \bigskip
    
    \begin{tabular}{| l | c c | c c |}
    
        \hline
        
        &
            \multicolumn{2}{c |}{\textbf{avg. classes per sample}}    &
            \multicolumn{2}{c |}{\textbf{avg. relations per sample}}  \\
        
        \multicolumn{1}{| c |}{\textbf{dataset}}  &
            \textbf{specified}                    &
            \textbf{inferable}                    &
            \textbf{specified}                    &
            \textbf{inferable}                    \\

        &
            (pos./neg.)  &
            (pos./neg.)  &
            (pos./neg.)  &
            (pos./neg.)  \\
        
        \hline
        
        \rowcolor{\RowColor}
        family trees    &  23 / ---   &  --- / 23     &  28 / ---   &  240 / 16,160  \\
        countries (S1)  &  --- / ---  &  238 / 478    &  820 / ---  &  20 / 49,261   \\
        \rowcolor{\RowColor}
        countries (S2)  &  --- / ---  &  238 / 478    &  782 / ---  &  39 / 49,279   \\
        countries (S3)  &  --- / ---  &  238 / 478    &  757 / ---  &  68 / 49,275   \\
        \rowcolor{\RowColor}
        DBpedia         &  --- / ---  &  183 / ---    &  642 / ---  &  156 / ---     \\
        Claros          &  --- / ---  &  1,499 / ---  &  518 / ---  &  17,072 / ---  \\
        \rowcolor{\RowColor}
        UMLS-reasoning  &  43 / 39    &  1,194 / ---  &  28 / 44    &  59 / 345      \\
        
        \hline
    
    \end{tabular}
    
    \caption{%
        This table breaks down the datasets that we used in our experiments according to the total
        numbers of samples, individuals, and triples that they contain.
        For triples, it lists both class memberships and relations specified as facts as well as
        such that are just inferable.
        The numbers of positive and negated triples are quoted separately.
        Notice that we trained and evaluated an RRN on 20 independent datasets for each countries
        task in order to obtain a larger amount of evaluation data.
        To that end, the number of training samples are average values over all these datasets,
        while the number of test samples was summed up over all of them.
    }
    \label{tab:datasets}
\end{table}

To assess the suggested approach, we trained and evaluated an RRN on five different datasets, two
out of which were artificially generated toy datasets, two were extracted from real-world databases,
and one was a generated dataset based on a real-world ontology.
Toy problems, generally, have the great advantage that it is immediately evident how difficult
certain kinds of inferences are, and thus provide us with a fairly good impression of the model's
abilities.
Nevertheless, evaluating an approach in a real-world setting is, of course, an indispensable measure
of performance, which is why we considered both kinds of data for our experiments.
Table~\ref{tab:datasets} provides a few summary statistics of the used datasets.
In this table as well as in the sequel, we use the terms specified or inferable classes and
relations as a short form of triples that describe class memberships and relations, respectively.
This means that \MyQuote{inferable classes}, for example, refers to triples that are inferable from
(but not given as facts in) a knowledge base, and that specify a class membership of an individual.
Notice that we always assume ontologies to be fixed, which is why there is no indication for
talking about specified or inferable class or relation types.
Notice that the stated counts of individuals, classes, and relations in Table~\ref{tab:datasets}
are average values per sample knowledge base.
Furthermore, triples that describe classes and relations, respectively, appear both in positive and
negative from.
For example, $\Triple{i}{\Member}{C_\ell}$ specifies the positive literal $C_\ell ( i )$, while
$\Triple{i}{\neg\Member}{C_\ell}$ describes a negative one $\neg C_\ell ( i )$.
Accordingly, Table~\ref{tab:datasets} provides separate counts for triples specifying positive and
negative literals, respectively, and in the sequel, we commonly refer to positive or negatives
facts and inferences, respectively, which shall be interpreted this way.

Each of the considered datasets is a collection of sample knowledge bases that share a common
ontology.
During training and evaluation, the model was provided with all the facts in the considered
samples, and had to compute predictions for both facts and inferences that were derivable from the
same, based on the used ontologies.
Since neural networks are notoriously data-hungry, there has been a recent interest in training
models on limited amounts of data.
Taking the same line, we confined all  our training sets to a total of 5000 training samples.


\subsection{Datasets}

\ToyDatasetFigure

We generated two toy datasets that pose reasoning tasks of varying difficulty with respect to family
trees, on the one hand, and countries, on the other.
In the family-trees dataset (Figure~\ref{fig:toy-datasets}a), samples describe pedigrees of
different sizes such that the only facts that are available in any of them are the genders of the
people involved as well as the immediate ancestors, that is, parent-of relations, among them.
Besides this, the used ontology specifies rules for reasoning about 28 different kinds of family
relations, ranging from \MyQuote{easy} inferences, such as fatherhood, to more elaborate ones,
like being a girl first cousin once removed.\footnote{%
    The term \MyQuote{girl first cousin once removed} refers to the daughter of someone's cousin.
}
What is particularly challenging about this dataset, however, is that seemingly small samples allow
for great numbers of inferences.
For instance, an actual sample that consists of 12 individuals with 12 parent-of relations specified
among them admits no less than 4032 inferences, both positive and negative.
Furthermore, the training data is highly skewed, as, for most of the relation types, only less than
three percent of all inferences are positive ones.
Appendix~\ref{app:family-trees} contains additional details of how we created this dataset.

The second toy dataset is based on the countries knowledge base \citep{Bouchard2015}, which
describes the adjacency of countries together with their locations in terms of regions and
subregions.
In every sample, some of the countries' regions and subregions, respectively, are not stated as
facts, but supposed to be inferred from the information that is provided about their neighborhoods
(Figure~\ref{fig:toy-datasets}b).
Following \citet{Nickel2016}, we constructed three different versions of the dataset, S1~(easy), S2,
and S3~(hard).
In contrast to the original work, however, we created an ontology that allows for reasoning not just
over countries' locations, but also about their neighborhood relations.
Furthermore, we introduced classes for countries, regions, and subregions, all of which have to be
inferred from the provided set of facts.
An interesting characteristic of the latter two versions of the dataset is that the sample knowledge
bases are constructed such that parts of the missing relations cannot be inferred by means of the
ontology at all.
The same is true for some of the class memberships in all three versions of the problem.
This challenges the model's ability to generalize beyond pure ontology reasoning.
Additional details about how we generated the countries datasets can be found in
Appendix~\ref{app:countries}.

To evaluate the RRN model on real-world data, we extracted datasets from two well-known knowledge
bases, DBpedia \citep{Bizer2009} and Claros.
The former of these represents part of the knowledge that is available in terms of natural
language on Wikipedia, and the latter is a formalization of a catalog of archaeologic artifacts,
which was created as part of the RDFox project \citep{Nenov2015}.
Since these datasets do not naturally separate into samples, we extracted sample knowledge bases
that are subgraphs of the original knowledge graphs, each of which contains a total of 200
individuals.
This way, it is possible to evaluate the model on sample knowledge graphs that it has not seen as
part of the training.
Appendix~\ref{app:real-world-data} provides additional details about how we extracted those samples.

While Claros makes use of a total of 33 classes and 77 relations, DBpedia employs a massive
vocabulary consisting of thousands of classes and relations, respectively.
To make the according experiments more computationally feasible, we restricted this to the 101
most frequent classes and those 518 relation types that allow for the greatest numbers of
inferences.
What is interesting about both of the considered knowledge bases is that neither of them contains
any specified, but only inferable class memberships, which is a very common characteristic of
real-world datasets.
Furthermore, the distribution of relations with respect to individuals differs quite heavily in
both of them.
While the branching factor in the knowledge graph specified by DBpedia varies rather smoothly,
Claros contains certain clusters of individuals that share very large numbers of relations,
whereas their vicinities are comparably sparse.
This is reflected by the fact that some sample knowledge bases consist of a few hundred specified
relations only, while others contain more than 100,000.

In addition to this, we created a dataset based on the Unified Medical Language System
(UMLS; \citealt{McCray2003}), an ontology that has been introduced for describing concepts from the
biomedical domain in a uniform way.
The UMLS ontology makes use of 127 classes and 53 relations, and thus, in contrast to DBpedia and
Claros, contains more classes than relations.
The UMLS is just an ontology, however, and does not specify any facts at all, which is why we
generated a dataset of 5000 sample knowledge bases, all of which make use of the UMLS ontology as a
training set.
Additional details about the used generation process are provided in
Appendix~\ref{app:real-world-data}.

Interestingly, the UMLS is commonly used as a benchmark dataset for methods of knowledge base
completion \citep{Kok2007}.
In this context, however, the UMLS is not interpreted as an ontology, but simply viewed as an
ordinary knowledge graph.
This is possible, as the structure of the ontology is defined in terms of triples, such as
$\Triple{Bacterium}{isa}{Organism}$, which may be interpreted as the facts that specify a knowledge
graph, and in the benchmark dataset, some of these triples are missing and thus have to be
predicted based on the remaining information.
In the sequel, we use the term \MyQuote{UMLS-reasoning} to refer to our own dataset, and set it
apart from the benchmark by \citep{Kok2007}, which is commonly just referred to as \MyQuote{UMLS}.

Finally, notice that we partitioned each of our datasets into pairwise disjoint sets for training
and testing, which ensures that the model actually learns to perform ontology reasoning as opposed
to just memorizing the data.
Also, we would like to point out that, unfortunately, \citet{Ebrahimi2018}, who cited our work,
mistakenly claimed that we trained and evaluated our models on one and the same data, which is
\emph{not} the case.


\subsection{Results}

\begin{table}
    \centering
    \footnotesize
    \setlength{\tabcolsep}{5pt}
    
    \begin{tabular}{| l | c c c c | c c c c |}
        \hline
                                                    &
        \multicolumn{4}{c |}{\textbf{accuracy on}}  &
        \multicolumn{4}{c |}{\textbf{F1 score on}}  \\
        \multicolumn{1}{| c |}{\textbf{dataset}}    &
        \textbf{spec.}                              &
        \textbf{inf.}                               &
        \textbf{spec.}                              &
        \textbf{inf.}                               &
        \textbf{spec.}                              &
        \textbf{inf.}                               &
        \textbf{spec.}                              &
        \textbf{inf.}                               \\
                                                    &
        \textbf{classes}                            &
        \textbf{classes}                            &
        \textbf{relations}                          &
        \textbf{relations}                          &
        \textbf{classes}                            &
        \textbf{classes}                            &
        \textbf{relations}                          &
        \textbf{relations}                          \\ \rowcolor{\RowColor}
        \hline
        family trees    &
            1.000       &
            1.000       &
            1.000       &
            0.999       &
            1.000       &
            1.000       &
            1.000       &
            0.976       \\
        countries (S1)  &
            ---         &
            1.000       &
            1.000       &
            0.999       &
            ---         &
            1.000       &
            1.000       &
            0.999       \\ \rowcolor{\RowColor}
        countries (S2)  &
            ---         &
            1.000       &
            0.999       &
            0.999       &
            ---         &
            1.000       &
            0.997       &
            0.929       \\
        countries (S3)  &
            ---         &
            1.000       &
            0.999       &
            0.999       &
            ---         &
            1.000       &
            0.996       &
            0.916       \\ \rowcolor{\RowColor}
        DBpedia         &
            ---         &
            0.998       &
            0.998       &
            0.989       &
            ---         &
            0.997       &
            0.998       &
            0.962       \\
        Claros          &
            ---         &
            0.999       &
            0.999       &
            0.996       &
            ---         &
            0.999       &
            0.999       &
            0.997       \\ \rowcolor{\RowColor}
        UMLS-reasoning  &
            0.989       &
            0.990       &
            0.997       &
            0.997       &
            0.969       &
            0.994       &
            0.996       &
            0.989       \\
        \hline
    \end{tabular}

    \caption{%
        This table summarizes our experimental results.
        Accuracy and F1 score are reported separately for class memberships and relations, and
        within each group, for those triples that describe specified knowledge, that is, facts, and
        those that represent inferable information.
    }
    \label{tab:results}
\end{table}

Table~\ref{tab:results} summarizes the results that have been achieved in our experiments.
For reports in much greater detail, we refer the reader to Appendix~\ref{app:family-trees-details}
and~\ref{app:countries-details}, which provide additional information about results on the toy
datasets.
The detailed results of our experiments with DBpedia, Claros, and UMLS-reasoning are too
comprehensive to be included in this paper, and are thus provided online at http://paho.at/rrn.
In addition to this, Appendix \ref{app:experimental-setup} provides a description of our
experimental setup.

There are a number of interesting aspects to notice across all the considered reasoning tasks.
First, we see that the RRN is able to effectively encode provided facts, about both classes and
relations, as these are predicted correctly from the created embeddings with an accuracy greater
than 98.9\%, and except for UMLS-reasoning, even with more than 99.8\%.
Furthermore, we observe that the model also learns to reason over classes with an equally high
accuracy of 99.8\%
on all datasets except UMLS-reasoning, for which we find an accuracy of 99.0\%.
This difference is not surprising, though, as the UMLS ontology specifies by far the largest
taxonomy of classes.
Therefore, separate class prediction layers might be required to achieve an even higher accuracy on
predicting inferable class memberships in this dataset.
For reasoning over relations, we find a slightly lower accuracy of 98.9\% on DBpedia, while
derivable relations in other datasets are predicted correctly in at least 99.6\% of all cases.
Again, however, this difference is not surprising, as DBpedia's is by far the most complex among the
ontologies that were used in our experiments, and might thus require a larger training dataset
in order to achieve an even higher accuracy.
Also, it was to be expected that the model would generally perform better in predicting inferable
classes than relations, since most of these are inferences depending on single triples only.

As stated in the detailed result tables, comparing specified with inferable triples reveals that,
throughout all datasets, there are imbalances with respect to positive and negative relations,
that is, the majority of specified triples is positive, while most inferable ones are negative.
Therefore, we would like to stress at this point that the RRN employs the same MLPs for predicting
both kinds of triples, and thus cannot \MyQuote{cheat} by leveraging those imbalances in the data.
Also, the great surplus of negative triples among all inferable ones explains why the reported F1
values are slightly lower than the accuracies quoted.

Earlier, we pointed out a few problems that symbolic methods of KRR commonly struggle with.
To evaluate whether the RRN suffers from these issues as well, we corrupted the test data of our
real-world datasets, DBpedia and Claros, and examined whether the models that have been trained as
part of our experiments are able to resolve the introduced flaws.
For the problem of missing information, we randomly removed one fact that could not be inferred by
means of symbolic reasoning from each sample, and checked whether the model was able to reconstruct
it correctly.
For DBpedia, this was the case for 33.8\% of all missing triples, and for Claros, 38.4\% were
predicted correctly.
Just like this, we tested the model's ability to resolve conflicts by randomly choosing one fact in
each test sample, and adding a negated version of the same as yet another fact.
For DBpedia, the RRN resolved 88.4\% of the introduced conflicts correctly, and for Claros, it were
even 96.2\%.
Most importantly, however, none of the total accuracies reported earlier dropped by more than 0.9
for either of the corrupted datasets.
This is remarkable, as both kinds of deficiencies pose serious problems for symbolic methods, and
conflicting information, in particular, prohibits the same entirely for most of the formalisms that
are used today.

All the query predictions of the RRN  are solely based on the embeddings that it generated for the
individuals in the respective datasets, which is why it is instructive to have a closer look at such
a set of embedding vectors.
Figure~\ref{fig:embeddings} provides a two-dimensional t-SNE projection of those embeddings for
the individuals in one sample of the countries dataset (S1).
In this visualization, dots represent individuals, and their colors indicate whether they
represent countries, regions, or subregions.
Furthermore, each country and each subregion is endowed with a colored shadow that reveals the
region that the respective individual is located in.
Without even looking at the labels provided, one can easily detect groups of countries that
belong to one and the same subregion.
Furthermore, most of the clusters representing subregions that are located in the same region are
neighbors in this visualization as well.
Finally, observe that the individuals that represent regions and subregions, respectively,
are mostly encoded as embeddings in confined regions of the embedding space.

\EmbeddingsFigure

Despite a considerable body of related work (cf. Section~\ref{sec:related-work}), hardly any of the
introduced models can be used to address the same exact problem setting that is considered in this
work, such that a direct comparison would not be fair in most cases.
There is one exception, however, and this is the Neural Theorem Prover
(NTP; \citealt{Rocktaschel2017}).
Although it has been introduced in the context of knowledge graph completion, the NTP is based on
the backward chaining algorithm that is used in Prolog, and can thus be used for logical reasoning.
Since the NTP model is rather involved, we refrain from providing a description here, and refer the
interested reader to the original paper by \citet{Rocktaschel2017} instead.
There are, however, two important aspects to mention.
First, it is possible for the NTP to make use of symbolic rules that are provided as part of the
input, which is why the model is suitable for ontology reasoning.
These rules are restricted to positive Horn rules, though, and thus cover just a fraction of the full
reasoning problem that we aim to solve in this paper.
Second, like most models for knowledge graph completion, the NTP is trained on the same knowledge
graph that is supposed to be completed, and thus cannot make use of the RRN's training scheme
described above.

As pointed out by \citet{Rocktaschel2017}, the NTP suffers from severe performance problems, which
is why we compared it to the RRN on our toy datasets only.
Furthermore, due to the restricted set of rules that can be provided to an NTP, we removed all those
rules from the used ontologies that are not supported by the model, and instead employed it for
reasoning under the CWA.
This way, it is possible to employ the NTP model for the considered reasoning problems on our toy
datasets.
Notice, however, that this is not the case in general.
Furthermore, we made use of the (empirically) best version of the NTP model, which is referred to
as NTP$\lambda$ in the original paper.

\begin{table}
    \centering
    \footnotesize
    \setlength{\tabcolsep}{5pt}
    
    \begin{tabular}{| c | c c c c |}
        \hline
        \multicolumn{1}{| c |}{\textbf{model}}                           &
        \multicolumn{4}{  c |}{\textbf{accuracy on inferable relation}}  \\
                        &
        family trees    &
        countries (S1)  &
        countries (S2)  &
        countries (S3)  \\ \rowcolor{\RowColor}
        \hline
        RRN        &
            0.999  &
            0.999  &
            0.999  &
            0.998  \\
        NTP        &
            0.971  &
            0.999  &
            0.890  &
            0.875  \\
        \hline
    \end{tabular}

    \caption{%
        The results of our experimental comparison of the RRN and the NTP.
        In this table, accuracies are stated with respect to predicting relations that are inferable
        relative to the used ontologies, and for the countries datasets, we considered the
        located-in relation only.
    }
    \label{tab:results-ntp}
\end{table}

Table~\ref{tab:results-ntp} summarizes our comparison of the RRN and the NTP on the family trees
and the different versions of the countries dataset.
For evaluating the models against each other, we considered inferable relations for computing
accuracies only, that is, we excluded facts as well as any triples about class memberships, as these
provide a better picture of a model's abilities.
On the family trees dataset, we evaluated both models with respect to all inferable relations, but
on the countries dataset, we confined ourselves to located-in relations only, as inferable relations
of other relation types are more or less trivial.
As stated in the table, the RRN, which achieves an accuracy of at least 99.8\% on all datasets,
outperforms the NTP in all the considered cases except version S1 of the countries data, for
which both models perform equally well.
On version S2 and S3 of the countries dataset, however, we observe a signification margin of more
than 10\% accuracy.

There is another interesting detail about the application of the NTP on the family trees dataset.
All facts about relations in the dataset are parent-of relations, which means that none of the
other predicates that appear in any inferable triples can be observed as facts.
As the NTP is always trained on the same knowledge base that it is evaluated on, however, it is not
able to learn anything about these relation types, and cannot even provide predictions for the same.
In contrast to this, the RRN can be trained on synthetically generated training data, and thus
learns to predict inferable relations almost perfectly.
To allow for comparing the RRN and the NTP on the family trees dataset anyway, we thus added half of
the inferable relations of each relation type as facts to the according sample knowledge bases.
Notice, however, that the RRN achieved the same accuracy of 99.9\% on the original version of the
family trees data, as stated in Table~\ref{tab:results}.


\subsection{Qualitative Evaluation}

Despite the convincing results, the RRN, like most models that are based on deep learning, remains
basically a black box for which it is not obvious how exactly predictions are computed.
Our experimental evaluation shows, however, that the model's computations are effectively guided by
the rules of the considered ontology, and we are planning to make  this aspect fully transparent in
future extensions of the RRN architecture.
In this section, we analyze further characteristics of the RRN model, and try to understand how it
depends on different design choices.


\bigskip
\noindent
\textbf{Update iterations.}
\
As described above, the number of update iterations is used to specify how often each fact in a
knowledge base is considered for updating the embeddings of the individuals involved in the same
(cf. Algorithm \ref{alg:embedding-generation}).
Intuitively, the reason why we have to perform multiple update iterations is that update steps are
local in the sense that they are based on a single fact together with the information that is
encoded in the embeddings of the individuals involved.
Often, however, ontologies contain rules that are based on multiple triples at the same time, for
example, if inferences depend on two individuals in a knowledge graph being connected by a path of
length greater than one, which is commonly referred to as multi-hop reasoning.
In a case like this, multiple update iterations are necessary, since more than two embeddings have
to be adjusted in order to transfer information between all individuals involved.

To verify this, consider Table~\ref{tab:update-iters}, which provides prediction accuracies for
inferable relations of different relation types in the family trees dataset.
The relations described in this table range from easy 1-hop inferences up to much more elaborate
5-hop reasoning, and the according accuracies were evaluated after each of 5 update iterations in
total.
Now, there are two interesting aspects to observe in this table.
First and foremost, we see that the number of update iterations that are necessary in order to
achieve the peak accuracies with respect to the different relation types is indeed correlated with
the according number of hops.
Even more interestingly, however, all accuracies are greater than 0.91 after the very first update
iteration already, and greater than 0.99 after the second one.
This suggests that the model learns to make effective use of multiple rules in the ontology at the
same time, and is not confined to drawing simple one-step inferences per update iteration.

\begin{table}
    \centering
    \footnotesize
    \setlength{\tabcolsep}{5pt}
    
    \begin{tabular}{| c | c c c c c |}
        \hline
        \textbf{update iters}                                               &
            \multicolumn{5}{c |}{\textbf{accuracy on inferable relations}}  \\
                                       &
            fatherOf                   &
            sisterOf                   &
            greatGrandsonOf            &
            girlCousinOf               &
            boyFirstCousinOnceRemoved  \\
                        &
            (1 hop)     &
            (2 hops)    &
            (3 hops)    &
            (4 hops)    &
            (5 hops)    \\ \rowcolor{\RowColor}
        \hline
        1          &
            0.992  &
            0.991  &
            0.921  &
            0.919  &
            0.940  \\
        2                   &
            \textbf{1.000}  &
            0.996           &
            0.995           &
            0.994           &
            0.992           \\ \rowcolor{\RowColor}
        3                   &
            1.000           &
            \textbf{0.997}  &
            0.999           &
            0.998           &
            0.999           \\
        4                   &
            1.000           &
            0.997           &
            \textbf{1.000}  &
            \textbf{0.999}  &
            0.999           \\ \rowcolor{\RowColor}
        5          &
            1.000  &
            0.997  &
            1.000  &
            0.999  &
            \textbf{1.000}  \\
        \hline
    \end{tabular}
    
    \caption{%
        The prediction accuracies for inferable relations with different predicates in the family
        trees dataset after different numbers of update iterations.
        Bold-faced values mark the lowest number of update iterations that allow for achieving the
        best possible accuracy for a relation type.
    }
    \label{tab:update-iters}
\end{table}


\bigskip
\noindent
\textbf{Number of layers.}
\
A noticeable property of the RRN's model architecture is that it, for both updates and predictions,
comprises one layer for each relation type that exists in the considered ontology, but just a
single one for class types altogether.
The intuition behind this is that inferences about class memberships tend to be \MyQuote{easier} in
the sense that they frequently depend on single triples only or can be drawn based solely on the
knowledge about the individual concerned.
Empirically, this is confirmed by the results presented in the previous section, as the predictive
accuracy on inferable classes is greater than or equal to the accuracy on inferable relations on all
considered datasets.
This raises the question of whether it is possible to reduce the number of layers that are used for
operations related to relations in a similar manner in order to reduce the total number of trainable
parameters.

It turns out, though, that this is not the case.
Table~\ref{tab:pred-layers} summarizes, for all datasets, how the accuracy of predicting inferable
relations changes if we use just a single compound prediction layer, like for predicting class
memberships.
To that end, we observe a significant drop for each of the datasets, which becomes more severe as
the number of relation type in the considered ontology grows.
On DBpedia, for example, the accuracy achieved with a single prediction layer is just slightly
better than random guessing.

\begin{table}
    \centering
    \footnotesize
    \setlength{\tabcolsep}{5pt}
    
    \begin{tabular}{| l | c c c c c |}
        \hline
        \textbf{dataset}    &
            family trees    &
            countries (S3)  &
            DBpedia         &
            Claros          &
            UMLS-reasoning  \\ \rowcolor{\RowColor}
        \hline
        \textbf{1 pred. layer per relation}  &
            0.999                            &
            0.999                            &
            0.997                            &
            0.989                            &
            0.996                            \\
        \textbf{1 pred. layer for all relations}  &
            0.778                                 &
            0.942                                 &
            0.563                                 &
            0.616                                 &
            0.644                                 \\
        \hline
    \end{tabular}
    
    \caption{%
        The accuracies for inferable relations that were achieved with one prediction layer per
        relation type on the one hand and a single prediction layer for all relations on the other.
    }
    \label{tab:pred-layers}
\end{table}




\section{Related Work}
\label{sec:related-work}


In this section, we provide a comparison to existing learning-based approaches to logical reasoning 
as well as to symbolic methods to logical reasoning.  In this context, we also discuss some 
general limitations of the presented RRN model.

\subsection{Learning-Based Logical Reasoning}
\label{sec:existing-work}

While there has been an increasing interest in the application of learning-based methods to various
kinds of logical reasoning in the last few years, ontology reasoning in particular has received just
modest attention.
The only published paper that we are aware of is by \citet{Makni2018}, who developed an approach to
RDFS reasoning contemporaneously with the work presented in this article.
To that end, \citeauthor{Makni2018} suggested an algorithm for mapping entire knowledge graphs to
embedding vectors, which they refer to as graph words.
Building on this, they use a BiGRU encoder-decoder architecture for translating the embedding of the
knowledge graph specified by a database to the embedding of the according inference graph.
Finally, the inference graph is reconstructed from the translated embedding, and predictions are
extracted from the same.
\citeauthor{Makni2018} evaluated their approach on two different datasets, a toy dataset that makes
use of the LUBM ontology \citep{Guo2005}, a very simple toy ontology, and real-world data that have
been extracted from DBpedia.
On the unimpaired versions of both datasets, they reported prediction accuracies of 0.98 and 0.87
for LUBM and DBpedia, respectively, which are well below the results that the RRN achieved on
the comparable datasets that we used in our own experiments.
Despite the fact that the RRN learns to perform ontology reasoning with a very high accuracy, it does not yet allow for providing justifications for
the predicted inferences,  such as via inference graphs like the model by \citet{Makni2018}.
The reason for this is that the RRN iteratively performs local updates of individual embeddings,
which does not easily allow for tracking global inference paths.
However, we are planning to extend the RRN architecture as part of future work in order to account
for this task.

Another very recent work by \citet{Ebrahimi2018} addresses reasoning over RDF knowledge bases via
an adapted version of end-to-end memory networks (MemN2N; \citealp{Sukhbaatar2015}).
To that end, \citeauthor{Ebrahimi2018} treat triples like text, and map the elements of any such
(that is, subject, predicate, and object) to what they call normalized embeddings, which are created
by first applying standard data preprocessing for logical reasoning, and then randomly mapping the
names of primitives of the considered logical language (that is, variables, constants, functions,
and predicates) to a predefined set of entity names.
The rationale behind this is that the model should learn that names of primitives do not matter,
and inferences depend on structural information about a knowledge base only.
After this, the embedded triples are placed in the memory of an adapted MemN2N, which computes a
prediction for a query that is also provided as an embedded triple.
To evaluate their approach, \citeauthor{Ebrahimi2018} extracted three different datasets, consisting
of sample knowledge bases with 1000 triples each, from various public sources.
They tested their model on a few different versions of each dataset, and reported 0.96 as the
highest accuracy that has been achieved on any of them.
In contrast to this, the RRN achieved an accuracy of more than 0.99 (taken over both classes and
relations) in our own experiments with real-world data. 
Furthermore, the model introduced by \citet{Ebrahimi2018} does not compute all inferences that may
be derived from a knowledge base, but provides predictions for single queries only. 
Differently from the approaches by \citet{Ebrahimi2018} and also \citet{Makni2018},
the RRN is trained relative to a fixed ontology, instead of reading
the same as part of the input.
This is a design choice that involves a bit of trade off, since it results in reasoning models of
unprecedented accuracy, as shown in our experiments, but requires us to train a new model from
scratch whenever a different ontology is considered.
We do believe, however, that this meets the requirements in practice, as ontologies are usually
considered as fixed in most use cases.


Apart from this, a lot of previous work addresses the combination of logic-based symbolic reasoning
and deep learning in some way, but is not related to ontology reasoning per se---notice that the
following is not an exhaustive account.
There exists a large body of previous work on neural-symbolic systems for learning and reasoning
\citep[see especially][]{Sun1997,Hammer2007,dAvila2012,dAvila2015},
which are classified into hybrid translational and hybrid functional systems.
While the former translate and extract symbolic representations into and from neural networks,
respectively, in which each symbolic sentence may be encoded in clearly localized neurons and in a
distributed way over all neurons, the latter couple a symbolic and a neural representation and
reasoning system \citep{Sun1997}.
However, a coupling of two systems in hybrid functional systems comes with a mismatch of data
learning capabilities, easy adaptability, and error tolerance between the two systems, while a
localized hybrid translational system essentially just moves this mismatch into a single neural
network.

Another line of research in neural-symbolic logical inference takes inspiration from the backward
chaining algorithm used in logic programming technologies such as Prolog.
More specifically, \citet{Rocktaschel2017} as well as \citet{Minervini2018} replace symbolic
unification with a differentiable computation on vector representations of symbols, using radial
basis function kernels.
The approach learns to place representations of similar symbols in close proximity in a vector
space, makes use of such similarities to prove queries, induces logical rules, and uses provided and
induced logical rules for multi-step reasoning.
It thus also combines logic-based reasoning with learning vector representations, but is intuitively
best described as a vector simulation of backward chaining. 
Differently from these approaches, the RRN is in particular not able to induce additional ontological rules. 
Closely related are logic tensor networks \citep{Serafini2016,Donadello2017}, which are different
from the approaches by \citet{Rocktaschel2017} and \citet{Minervini2018} in that they fully ground
first-order logic rules and also support function terms.
Other neural-symbolic approaches focus on first-order inference, but do not learn subsymbolic vector
representations from training facts in a knowledge base, for example, CLIP++ \citep{Franca2014},
lifted relational neural networks \citep{Sourek2015}, Neural Prolog \citep{Ding1995},
SHRUTI \citep{Shastri1992}, and TensorLog \citep{Cohen2016,Cohen2017}.
Further related earlier proposals for neural-symbolic networks for logical inference are limited to
propositional rules, for example, C-IL~2~P \citep{dAvila1999}, EBL-ANN \citep{Shavlik1989}, and
KBANN \citep{Towell1994}, including the recent ones by \citet{Bowman2014,Bowman2015}, which present
successful approaches to simple propositional forms of logical reasoning.
These approaches are based on recursive neural tensor networks, and consider the binary logical
relationships entailment, equivalence, disjointness, partition, cover, and independence on
elementary propositional events. 

Another line of work
\citep{Rocktaschel2015,Demeester2016,Hu2016,Vendrov2016,Minervini2017,Minervini2017a}
regularizes distributed representations via domain-specific rules, but these approaches often
support a restricted subset of first-order logic only.
In particular, 
\citet{Rocktaschel2015} and \citet{Demeester2016} incorporate
implication rules relative to ground terms into distributed representations for natural language
processing, while \citet{Hu2016} incorporate ground instances of first-order rules into deep
learning, projected to the ground terms of the learned data in each minibatch.
Even with such restrictions, in experimental results in sentiment analysis and in named-entity
recognition, the approach achieves (with a few intuitive rules) substantial improvements and
state-of-the-art results to previous best-performing systems.
Closely related are some extensions of knowledge-base completion that additionally consider
non-factual symbolic knowledge to act as constraints to the learning process
\citep{Diligenti2012,Nickel2012}.
In particular, \citet{Diligenti2012} investigate a bridge between logic and kernel machines, and
use non-factual symbolic knowledge as constraints in the second step of a two-stage process, where
learning is done in the first one.
\citet{Minervini2017} account for very simple relationships between different relation types, such
as equivalence or inverse predicates, by making use of appropriate regularization terms.
Closely related to this, \citet{Xu2018} introduced the semantic loss, which is another type of
regularization term that allows for enforcing constraints that can be expressed by means of
propositional logic.
Another interesting approach 
by \citet{Minervini2017a} 
incorporates
background knowledge specified as Horn clauses into knowledge base completion by means of
adversarial training \citep{Goodfellow2014}.
Note that, in general, knowledge base completion \citep{Socher2013,Trouillon2017}, or link
prediction in knowledge bases, which is the problem of predicting non-existing facts in a knowledge
base consisting of a finite set of facts, differs from logical inference relative to a knowledge
base, as it is generally missing logical knowledge beyond simple facts.

Finally, notice that within the deep-learning literature, the term \MyQuote{reasoning} is
often used informally, that is, without reference to any kind of symbolic logic at
all \citep[e.g.,][]{Socher2013,Weston2015,Henaff2017,Santoro2017}.


\subsection{Symbolic Methods of Logical Reasoning}

One important question, which has been addressed briefly in the introduction already, is 
how to motivate the 
neural approach to ontology reasoning in the first place, and how the RRN
relates to purely symbolic methods of logical reasoning.

First, the presented neural approach to ontology reasoning 
is a step towards answering the wide open problem of 
how to combine deep learning technologies with symbolic methods for logical reasoning, 
which is  commonly regarded as a prerequisite for further substantial progress in AI. Implementing symbolic ontology reasoning with neural networks of very high accuracy
in some sense bridges the gap between neural and symbolic methods, and offers new ways of providing
machine learning models with reasoning capabilities that have previously been reserved for symbolic
methods only, which opens up interesting new opportunities.
From a machine-learning perspective, the RRN can be considered as a method of knowledge-graph
embedding \citep{Wang2017} that produces semantically meaningful embeddings of the individuals
in a knowledge graph.
These, in turn, may serve as input to models that are used for learning downstream tasks, and thus
allows for leveraging symbolic background knowledge in learning deep neural networks 
(and thus for knowledge transfer and for learning from smaller amounts of data) and explainable symbolic inference in computing according predictions towards a better explainability of the learned neural systems.

Second, the neural approach to ontology reasoning is useful in its own right as an alternative to 
symbolic methods to logical reasoning. Even though it does not allow for fully accurate logical reasoning, 
it paves the way for highly scalable implementations of nearly accurate approximate logical reasoning 
via parallel computations on GPUs. Such implementations may not be used for safety-critical applications
(such as for verifying the control of nuclear power plants), but it may be sufficiently accurate for 
many other applications where full accuracy is not required (such as question answering over the web). 

In the same vein, one major issue that many symbolic approaches, including all reasoning formalisms rooted in
classical logic, suffer from is conflicting information.
In practice, however, information is frequently imperfect, which is why conflicts inevitably have to
be dealt with in many use cases that symbolic reasoning is applied to.
While many reasoning methods simply do not work in any such case, our experiments with the RRN
suggest that the model is able to effectively resolve conflicts and thus better suited for applying
logical reasoning in a real-world scenario.
Even though there exist formalisms for reasoning over inconsistent
knowledge bases that are also able to resolve
conflicts, such as inconsistency-tolerant reasoning \citep{LLRRS10} and paraconsistent logics \citep{Middelburg2011},  these approaches are generally quite
limited in practice, since there is a price to be paid in terms of computational complexity.
Apart from confined special cases, theoretically powerful methods are usually computationally harder than
their counterparts for the consistent case (see, e.g., \citealp{LMPS15}), whereas the RRN allows for reasoning in linear time, irrespective of any conflicts that
may or may not exist in the considered knowledge base.

Finally, another challenge that is commonly encountered in practice is missing information, that is, details
that are neither specified as facts in the considered knowledge base nor inferable from them via symbolic reasoning.
Strictly speaking, recovering such missing pieces is a prediction rather than a reasoning task, and
hence usually not considered in the context of symbolic reasoning.
This is not in line with the requirements that are typically faced in practice, though, as we
frequently seek to do both, compute predictions and perform reasoning.
Again, however, our experiments indicate that this is exactly what the RRN does.
To that end, we observed that the model is able to provide sensible predictions for any missing
details that are compatible with the considered set of facts relative to the used ontology in many
cases, and that these, in turn, also affect inferences computed by the model.





\section{Conclusion}

The results of this work show that the RRN model is able to learn to effectively reason over diverse
ontological knowledge bases, and, in doing so, is the first one to achieve an accuracy that is very
close to the yet unattainable accuracy of symbolic methods, while being distinctly more robust.
Notice further that this was possible without the necessity of any kind of external memory, as it is
used by many state-of-the-art models of deep learning, such as the differential neural
computer~(DNC; \citealp{Graves2016}) or memory networks \citep{Weston2015}.
This paves the way for applying comprehensive logical reasoning to a range of important problems
that logic-based symbolic methods are, in general, hard to apply to.
In addition to standard ontology reasoning tasks with missing or conflicting information, these
include tasks such as understanding natural language, interpreting visual inputs, or even
autonomous driving, which is highly demanding in terms of reasoning about an agent's surrounding.
Furthermore, training an RRN relative to a suitable ontology is a simple and at the same time very
powerful way to provide a model with domain knowledge, which is difficult to achieve with most of
the state-of-the-art methods.

The RRN is among the very first deep-learning-based approaches to comprehensive ontology reasoning,
which is why it is hard to compare to the state-of-the-art of machine learning models for reasoning,
whose architectures do not allow for performing the same kind of inferences per se.
Still, it is interesting to observe the performance of related approaches on learning tasks that are
included as part of the reasoning problems presented in this work.
For instance, the DNC was previously evaluated on the problem of determining relationships based on
family trees \citep{Graves2016}, very much like the reasoning problem presented earlier.
In doing so, it achieved an average accuracy of 81.8\% on predicting four-step family relations,
while the RRN predicted more than 99.9\% of those relations correctly.
Another interesting comparison is with the recently introduced Neural Theorem
Prover~(NTP; \citealp{Rocktaschel2017}), which can indeed be used for simple ontology reasoning.
As shown above, however, the RRN outperforms the NTP throughout all datasets, sometimes by more than
10\% accuracy.
The substantial margins observed do suggest that the RRN is able to grasp reasoning concepts that
other state-of-the-art methods struggle with.
The reason for this, we believe, is that our holistic approach to ontology reasoning adds a lot more
structure to the considered data than other narrower prediction tasks, and hence makes it much
easier for a model to \MyQuote{understand} the same.
Furthermore, ontology reasoning, in many cases, naturally allows for breaking down complex
inferences into several easier reasoning tasks, which, in some sense, provide guidance for learning
to draw more elaborate conclusions.

Besides this, as discussed in Section 5, the RRN is superior to machine learning models for ontology
reasoning in particular as well.
This is most likely due to the fact that each RRN is trained relative to a fixed ontology, which
allows for adjusting its architecture appropriately, while other existing approaches consider the
specification of the same as part of the input.
Hence, there seems to be a certain tradeoff between generality, and thus independence of the
considered ontology, on the one hand, and reasoning accuracy, on the other.
In practice, however, ontologies are usually considered to be fixed, which is why a higher accuracy
is, in general, more desirable than a reasoning model that is not tied to a single ontology.

Finally, there is one subject that we raised just incidentally at the very beginning of this
article.
Despite the fact that the human brain served as a major source of inspiration for the
development of artificial neural networks \citep{Hassabis2017},
most network architectures that are used for machine learning today lack biological plausibility.
However, while many mechanics of the human neurology remain uncharted, there exist a number of
aspects that are considered to be confirmed by now, and some of them allow for drawing interesting
parallels to the RRN model.
For instance, there is a broad consensus that reasoning in the human brain is not realized like
logic-based symbolic reasoning, neither conceptually nor computationally.
Instead, it is believed that our brain maintains a probabilistic cognitive model of the
world \citep{Oaksford2007}, which provides a base for any kind of thought or action.
To that end, the storage of semantic memory is organized in a distributed way, and information is
encoded by strengthening synaptic connections among some neurons, while others are left to be
strangers to one another \citep{Martin2001}.
This is similar to how the RRN adjusts individual embeddings, which could be considered as groups of
neurons, whenever a new piece of information arrives.
Both absolute as well as relative positions of such vectors in the embedding space determine what is
believed to be true about the world, which is why the adjustment of those may be interpreted as
strengthening certain connections between neurons that store different pieces of information in a
distributed way.
Another interesting point is that the RRN, unlike other recent models, conducts logical reasoning
without any kind of external memory, which is not believed to happen in the human brain either.
Instead, reasoning is entangled with the generation of individual embeddings in the presented model,
and thus part of encoding information in a distributed manner.

At the bottom line, the RRN embodies a surprising pairing of results, as it
(i) learns to conduct highly accurate reasoning in a logic-based sense,
(ii) is able to work with complex real-world knowledge bases, and
(iii) is biologically plausible in a number of ways.

Despite the fact that an RRN is effectively guided by the ontology that it has been trained on, it
cannot provide justifications for predictions yet.
As an important future work, we are thus currently extending the presented architecture such that it allows for
inducing 
rules and 
explanations, such as inference graphs, alongside computed predictions, hence making the
RRN a fully interpretable neural reasoner.
Other interesting topics for future research include the generalization of the presented approach to more expressive ontology languages, such as those involving existential rules like the Datalog$^\pm$ family  \citep{DBLP:journals/ws/CaliGL12},  
and to other forms of logical reasoning, such as default reasoning \citep{eiter2000default}  
and~argumentation~\citep{bench2007argumentation}.


\acks{%
    This work was supported by the UK Engineering and Physical Sciences Research Council~(EPSRC)
    under the grants EP/J008346/1, EP/R013667/1, EP/L012138/1, and 
    EP/M0\-2526\-8/1, as well as the Alan Turing Institute under the EPSRC grant EP/N510129/1.
    Fur\-ther\-more, Patrick is supported by the EPSRC under the grant OUCL/2016/PH and the
    Ox\-ford-DeepMind Graduate Scholarship under the grant \mbox{GAF1617\_OGSMF-DMCS\_1036172}.
    We also acknowledge the use of the EPSRC-funded Tier 2 facility JADE (EP/P020275/1).
}


\appendix

    
\section{Experimental Setup}
\label{app:experimental-setup}

\begin{table}
    \centering
    \footnotesize
    
    \begin{tabular}{| l | c c c c c c |}
    	\hline
        \multicolumn{1}{|c|}{\textbf{dataset}} &
            \textbf{embedding}                 &
            \textbf{update}                    &
            \textbf{initial}                   &
            \textbf{batch}                     &
            \textbf{weight}                    &
            \textbf{MLP hidden}                \\
                                               &
            \textbf{size}                      &
            \textbf{iterations}                &
            \textbf{learning rate}             &
            \textbf{size}                      &
            \textbf{decay} $\lambda$           &
            \textbf{layers}                    \\ \rowcolor{\RowColor}
        \hline
        family trees   & 100 & 7 & 0.001 & 1 & $10^{-6}$ & 1 \\
        countries (S1) & 100 & 2 & 0.001 & 1 & $10^{-6}$ & 1 \\ \rowcolor{\RowColor}
        countries (S2) & 200 & 5 & 0.001 & 1 & $10^{-6}$ & 1 \\
        countries (S3) & 200 & 5 & 0.001 & 1 & $10^{-6}$ & 1 \\ \rowcolor{\RowColor}
        DBpedia        & 300 & 5 & 0.001 & 1 & $10^{-6}$ & 1 \\
        Claros         & 200 & 5 & 0.001 & 1 & $10^{-6}$ & 1 \\ \rowcolor{\RowColor}
        UMLS-reasoning & 300 & 5 & 0.001 & 1 & $10^{-6}$ & 1 \\
        \hline
    \end{tabular}
    
    \caption{%
        A summary of the hyperparameters that were used to achieve the results presented in this
        work.
    }
    \label{tab:hyperparameters}
\end{table}

As part of the conducted experiments, we performed a grid search in order to determine an
appropriate set of hyperparameters, all of which are reported in Table~\ref{tab:hyperparameters}.
Interestingly, the RRN seems to be broadly task-agnostic, as similar values worked well for all
the considered datasets.
Merely the size of the individual embeddings as well as the number of update iterations had to be
adjusted based on the respective reasoning task.
Furthermore, there was no need to manually create mini-batches of training data, as the single
training samples contained numerous triples for each class and relation type already.
All our models were trained by means of Adam \citep{Kingma2015} with an initial learning rate of
0.001, $\beta_1$ set to 0.9, and $\beta_2$ set to 0.999.
Furthermore, all MLPs had a single hidden layer of ReLU units and sigmoid units on
the output layers.
The sizes of the hidden layers were set to the average of input and output sizes for all of them.
To prevent overfitting, we employed weight decay, which means that we added the term
$\lambda \Norm{\Parameters}$ to the computed loss, where $\Parameters$ represents a vector of all
model parameters, and $\lambda \in \R_{\geq 0}$ is a hyperparameter.
The actual training loss was chosen to be a standard cross-entropy loss.

We trained our models for about two days on the toy datasets, and about seven days on the real-world
datasets.
However, for our experiments, we used a straightforward CPU-only implementation of the RRN model,
which did not make use of any optimization or parallelization strategies.
Therefore, it should be possible to reduce the duration of the model training by several orders of
magnitude by making use of different ways to optimize the implementation of the model, as outlined
in Section~\ref{sec:complexity}, as well as parallelizing computations over (multiple) GPUs.
Generating embeddings for a single sample knowledge base took, on average, about 10 seconds for the
toy datasets as well as Claros, and about 20 seconds for samples from DBpedia.
Computing predictions for single queries took less than a tenth of a second for all datasets.
Once again, however, it should be possible to reduce these durations significantly by optimizing the
code.

    
\section{Generation of Family-Tree Data}
\label{app:family-trees}

For the family-tree dataset, we generated training samples incrementally.
For each of these samples, starting from a tree that consisted of a single person only, we
randomly selected an existing person, and, again randomly, added either a child or, if possible,
a parent to the current family tree.
This process was repeated until either the maximum tree size, which we set to 26 people, was
reached or the sample generation was stopped randomly, which happened with a probability of
0.02 whenever another person was added to a tree.
As part of the data generation, we constrained sample family trees to a maximum depth of 5 as
well as a maximum branching factor of 5.
Furthermore, we ensured that the created dataset does not contain isomorphic family trees, such that
it is not possible for a model to just overfit the training data.
Finally, we computed all possible inferences for each of the created samples via symbolic
reasoning.


\section{Generation of Countries Data}
\label{app:countries}

Following the instructions in \citet{Nickel2016}, we generated three datasets of varying difficulty,
and endowed them with the formal ontology:
\begin{itemize}
    \item[\textbf{S1)}] This is the easiest version of the considered reasoning problem.
                        Here, some of the countries' regions are missing, and thus have
                        to be inferred from the knowledge about the subregions that those
                        countries belong to, on the one hand, and the regions that the subregions
                        are part of, on the other.
    \item[\textbf{S2)}] In this variation of the problem, there are countries that have neither
                        a region nor a subregion specified, and both of them have to be inferred
                        from what is known about their neighborhoods.
                        Notice that missing details in the test and evaluation set cannot be
                        completed  via ontology reasoning only, since countries that are
                        situated at the border of a territory possess neighbors with differing
                        regions and subregions, respectively.
    \item[\textbf{S3)}] This is the hardest of the tasks considered.
                        To that end, in addition to the problem described by version S2, the
                        neighbors of those countries whose regions and subregions are missing do not
                        have a region specified either.
                        Again, problems of this kind cannot be solved solely by ontology reasoning.
    \end{itemize}

While the original task considered reasoning about countries' locations only, we augmented the
problem scenario with three classes that represent the types of individuals that occur, that is,
countries, regions, and subregions.
Irrespective of the considered version of the problem, none of these are ever provided as facts, and
thus always have to be predicted as part of the reasoning problem.
To that end, just like it is the case for some of the relations, classes cannot always be inferred
via the ontology.
This is the case, for example, if a region does not have any subregions at all, which means that we
cannot leverage the transitivity of the located-in relation.

Every time we created a sample, we randomly selected 20 countries, and removed information about
them, as required by the considered problem setting.
In doing so, we ensured that every test country has at least one neighbor that is not part of the
test set itself.
For evaluating a trained model, we created two samples with such hold-out sets of countries for
evaluation and testing, respectively, with two distinct sets of countries used as test individuals.
To ensure that the model does not just overfit the data, we removed those 40 test countries from the
knowledge base before we generated 5000 samples as training data.
In addition to this, we included only those inferences in the evaluation and test sample that
concerned at least one of the test individuals, again to make sure that the model cannot just
overfit the training data, the only exception being regions and subregions, for which we included
class predictions as well.
Unlike this, training samples contain all inferences~whatsoever.

Finally, in order to obtain a larger amount of test data, we generated 20 independent datasets for
each of the three versions outlined above.
For every one of those, we trained an RRN on the training data, and evaluated the model on the
according test sample.
The results reported in this work summarize the outcomes of all these evaluation runs.

    
\section{Preparation of Real-world Data}
\label{app:real-world-data}

For both considered real-world databases, DBpedia and Claros, we extracted training samples
that are subgraphs of the knowledge graphs that are defined implicitly by the original data.
To that end, the single training samples were created by running breadth-first search, starting
from a randomly selected individual, on the knowledge graph that was specified by the facts in the
respective database until a total of 200 individuals was discovered.
Subsequently, all specified triples that concerned only those extracted individuals were considered
as the facts of the created sample knowledge base, and their inferences were computed via symbolic
reasoning.
Like most of the knowledge bases that are encountered in practice, both DBpedia and Claros consist
of positive facts and inferences only.
Therefore, we made use of the so-called local closed-world assumption \citep{Dong2014}, and
augmented the data with generated negative inferences that were created by randomly corrupting
either the subject or the object of existing positive triples.
This step is crucial, since the model would learn to blindly predict any queried inference to be
true, otherwise.
More precisely, we generated exactly one negative inference for each positive inference that exists
in the data by corrupting each of these positive inferences exactly once.
In doing so, we ensured, by means of symbolic reasoning, that the created triples do not introduce
inconsistencies, that is, conflicts, in the created sample knowledge bases.
Furthermore, whenever we generated a corrupted triple, we ensured that the same has not been added
to the dataset before.
For the evaluation and test data, we generated fixed sets of such negative inferences.
For the training data, however, new negative triples were generated on-the-fly in each training
iteration.

For the UMLS-reasoning dataset, we first created a Datalog program that implements that UMLS
ontology.
After this, we generated sample knowledge bases by means of an iterative procedure.
For each sample, we started from an empty knowledge base that contained 75 individuals, but no facts
about them.
Then, we randomly sampled a candidate fact, evaluated whether it was consistent with the current
state of knowledge relative to the ontology, and added it to the knowledge base, if possible.
In the sampling step, we first decided randomly whether to add a class membership or a relation,
both attributed a probability of 50\%, and then sampled the remaining parts of the according triple.
If an inconsistent fact was generated, then we created a new fact up to a total of 10 times, and if
sampling a consistent fact failed 10 times in a row, then we stopped the procedure for the current
sample knowledge base.
Otherwise, this step was repeated until a maximum of 200 triples had been added to a sample
knowledge base, and the remaining individuals that were not included in any fact got pruned again from
the same.
In addition to this, the generation of a sample knowledge base was stopped earlier with a probability of
0.5\% every time after a fact was added.
Finally, we computed all possible inferences for each sample knowledge base by means of symbolic
reasoning with the created Datalog program.

Like for the family trees data, we ensured that the created datasets do not contain isomorphic
sample knowledge graphs, such that it is not possible for a model to just overfit the training data.


\section{Data Availability}

All the datasets that have been used in our experiments are available from
https://paho.at/rrn.
Furthermore, the code that we used to generate our toy datasets, including the employed formal
ontologies, is available as open source from
https://github.com/phohenecker/family-tree-data-gen
and
https://github.com/phohenecker/country-data-gen,
respectively.


\section{Detailed Results for the Family Trees Dataset}
\label{app:family-trees-details}

In this section, we present detailed results of our experiments with the family-trees dataset.

\begin{center}

    \ResultTable{Results for Specified Classes}{ \rowcolor{\RowColor}
        0 &
        female &
            1.000 &
            1.000 &
            1.000 &
            1.000 &
            --- &
            5732 / 0 \\
        1 &
        male &
            1.000 &
            1.000 &
            1.000 &
            1.000 &
            --- &
            5702 / 0 \\
    }{
        \multicolumn{2}{c |}{1.000} &
        \multicolumn{2}{c |}{1.000} &
        \multicolumn{2}{c |}{1.000} \\
    }
    
    \par
    \medskip
    
    \ResultTable{Results for Inferable Classes}{ \rowcolor{\RowColor}
        0 &
        female &
            --- &
            --- &
            1.000 &
            --- &
            1.000 &
            0 / 5702 \\
        1 &
        male &
            --- &
            --- &
            1.000 &
            --- &
            1.000 &
            0 / 5732 \\
    }{
        \multicolumn{2}{c |}{---} &
        \multicolumn{2}{c |}{---} &
        \multicolumn{2}{c |}{1.000} \\
    }
    
    \par
    \medskip
    
    \ResultTable{Results for Specified Relations}{ \rowcolor{\RowColor}
        23 &
        parentOf &
            1.000 &
            1.000 &
            1.000 &
            1.000 &
            --- &
            28232 / 0 \\
    }{
        \multicolumn{2}{c |}{1.000} &
        \multicolumn{2}{c |}{1.000} &
        \multicolumn{2}{c |}{1.000} \\
    }
    
    \par
    \medskip
    
    \ResultTable{Results for Inferable Relations}{ \rowcolor{\RowColor}
        0 &
        auntOf &
            0.983 &
            0.998 &
            0.999 &
            0.998 &
            0.999 &
            8268 / 556484 \\
        1 &
        boyCousinOf &
            0.982 &
            0.999 &
            1.000 &
            0.993 &
            1.000 &
            5806 / 558946 \\ \rowcolor{\RowColor}
        2 &
        boyFirstCousinOnceRemovedOf &
            0.967 &
            0.999 &
            1.000 &
            0.999 &
            1.000 &
            2554 / 562198 \\
        3 &
        boySecondCousinOf &
            0.769 &
            0.989 &
            0.999 &
            1.000 &
            0.999 &
            958 / 563794 \\ \rowcolor{\RowColor}
        4 &
        brotherOf &
            0.908 &
            0.995 &
            0.996 &
            0.999 &
            0.996 &
            10388 / 554364 \\
        5 &
        daughterOf &
            0.991 &
            0.999 &
            1.000 &
            1.000 &
            1.000 &
            14116 / 550636 \\ \rowcolor{\RowColor}
        6 &
        fatherOf &
            0.992 &
            0.999 &
            1.000 &
            1.000 &
            1.000 &
            14116 / 550636 \\
        7 &
        girlCousinOf &
            0.966 &
            0.998 &
            0.999 &
            0.994 &
            0.999 &
            5586 / 559166 \\ \rowcolor{\RowColor}
        8 &
        girlFirstCousinOnceRemovedOf &
            0.944 &
            0.999 &
            0.999 &
            0.998 &
            0.999 &
            2522 / 562230 \\
        9 &
        girlSecondCousinOf &
            0.879 &
            0.995 &
            0.999 &
            0.998 &
            0.999 &
            1050 / 563702 \\ \rowcolor{\RowColor}
        10 &
        granddaughterOf &
            0.990 &
            1.000 &
            1.000 &
            1.000 &
            1.000 &
            13374 / 551378 \\
        11 &
        grandfatherOf &
            0.989 &
            0.999 &
            0.999 &
            1.000 &
            0.999 &
            13374 / 551378 \\ \rowcolor{\RowColor}
        12 &
        grandmotherOf &
            0.989 &
            0.999 &
            0.999 &
            1.000 &
            0.999 &
            13374 / 551378 \\
        13 &
        grandsonOf &
            0.993 &
            0.999 &
            1.000 &
            0.999 &
            1.000 &
            13374 / 551378 \\ \rowcolor{\RowColor}
        14 &
        greatAuntOf &
            0.979 &
            0.999 &
            1.000 &
            0.998 &
            1.000 &
            4840 / 559912 \\
        15 &
        greatGranddaughterOf &
            0.990 &
            1.000 &
            1.000 &
            1.000 &
            1.000 &
            9692 / 555060 \\ \rowcolor{\RowColor}
        16 &
        greatGrandfatherOf &
            0.989 &
            0.999 &
            1.000 &
            1.000 &
            1.000 &
            9692 / 555060 \\
        17 &
        greatGrandmotherOf &
            0.989 &
            1.000 &
            1.000 &
            1.000 &
            1.000 &
            9692 / 555060 \\ \rowcolor{\RowColor}
        18 &
        greatGrandsonOf &
            0.988 &
            1.000 &
            1.000 &
            1.000 &
            1.000 &
            9692 / 555060 \\
        19 &
        greatUncleOf &
            0.972 &
            0.999 &
            0.999 &
            1.000 &
            0.999 &
            4922 / 559830 \\ \rowcolor{\RowColor}
        20 &
        motherOf &
            0.989 &
            0.999 &
            0.999 &
            1.000 &
            0.999 &
            14116 / 550636 \\
        21 &
        nephewOf &
            0.972 &
            0.999 &
            0.999 &
            0.999 &
            0.999 &
            8450 / 556302 \\ \rowcolor{\RowColor}
        22 &
        nieceOf &
            0.976 &
            0.998 &
            0.999 &
            0.997 &
            0.999 &
            8376 / 556376 \\
        23 &
        parentOf &
            --- &
            --- &
            0.999 &
            --- &
            0.999 &
            0 / 536520 \\ \rowcolor{\RowColor}
        24 &
        secondAuntOf &
            0.958 &
            0.998 &
            1.000 &
            0.995 &
            1.000 &
            2518 / 562234 \\
        25 &
        secondUncleOf &
            0.946 &
            0.999 &
            0.999 &
            0.997 &
            0.999 &
            2558 / 562194 \\ \rowcolor{\RowColor}
        26 &
        sisterOf &
            0.922 &
            0.996 &
            0.997 &
            0.999 &
            0.997 &
            10072 / 554680 \\
        27 &
        sonOf &
            0.992 &
            0.999 &
            1.000 &
            1.000 &
            1.000 &
            14116 / 550636 \\ \rowcolor{\RowColor}
        28 &
        uncleOf &
            0.980 &
            0.999 &
            0.999 &
            0.998 &
            0.999 &
            8558 / 556194 \\
    }{
        \multicolumn{2}{c |}{0.976} &
        \multicolumn{2}{c |}{0.998} &
        \multicolumn{2}{c |}{0.999} \\
    }

\end{center}


\section{Detailed Results for the Countries Dataset}
\label{app:countries-details}

In this section, we present detailed results of our experiments with the different versions of the
countries dataset.


\subsection{Countries S1}

\medskip

\begin{center}

    \ResultTable{Results for Inferable Classes}{ \rowcolor{\RowColor}
        0             &
        country       &
            1.000     &
            1.000     &
            1.000     &
            1.000     &
            1.000     &
            400 / 580 \\
        1             &
        region        &
            1.000     &
            1.000     &
            1.000     &
            1.000     &
            1.000     &
            120 / 860 \\ \rowcolor{\RowColor}
        2             &
        subregion     &
            1.000     &
            1.000     &
            1.000     &
            1.000     &
            1.000     &
            460 / 520 \\
    }{
        \multicolumn{2}{c |}{1.000} &
        \multicolumn{2}{c |}{1.000} &
        \multicolumn{2}{c |}{1.000} \\
    }
    
    \par
    \medskip
    
    \ResultTable{Results for Specified Relations}{ \rowcolor{\RowColor}
        0             &
        locatedIn     &
            1.000     &
            1.000     &
            1.000     &
            1.000     &
            ---       &
            9160 / 0  \\
        1             &
        neighborOf    &
            1.000     &
            1.000     &
            1.000     &
            1.000     &
            ---       &
            10114 / 0 \\
    }{
        \multicolumn{2}{c |}{1.000} &
        \multicolumn{2}{c |}{1.000} &
        \multicolumn{2}{c |}{1.000} \\
    }
    
    \par
    \medskip
    
    \ResultTable{Results for Inferable Relations}{ \rowcolor{\RowColor}
        0               &
        locatedIn       &
            1.000       &
            1.000       &
            1.000       &
            1.000       &
            1.000       &
            400 / 10800 \\
        1               &
        neighborOf      &
            ---         &
            ---         &
            0.999       &
            ---         &
            0.999       &
            0 / 172904  \\
    }{
        \multicolumn{2}{c |}{0.999} &
        \multicolumn{2}{c |}{1.000} &
        \multicolumn{2}{c |}{0.999} \\
    }

\end{center}


\subsection{Countries S2}

\medskip

\begin{center}

    \ResultTable{Results for Inferable Classes}{ \rowcolor{\RowColor}
        0             &
        country       &
            1.000     &
            1.000     &
            1.000     &
            1.000     &
            1.000     &
            400 / 580 \\
        1             &
        region        &
            1.000     &
            1.000     &
            1.000     &
            1.000     &
            1.000     &
            120 / 860 \\ \rowcolor{\RowColor}
        2             &
        subregion     &
            1.000     &
            1.000     &
            1.000     &
            1.000     &
            1.000     &
            460 / 520 \\
    }{
        \multicolumn{2}{c |}{1.000} &
        \multicolumn{2}{c |}{1.000} &
        \multicolumn{2}{c |}{1.000} \\
    }
    
    \par
    \medskip
    
    \ResultTable{Results for Specified Relations}{ \rowcolor{\RowColor}
        0 &
        locatedIn &
            0.996 &
            0.999 &
            0.995 &
            0.995 &
            --- &
            8760 / 0 \\
        1 &
        neighborOf &
            0.999 &
            0.999 &
            0.999 &
            0.999 &
            --- &
            9888 / 0 \\
    }{
        \multicolumn{2}{c |}{0.997} &
        \multicolumn{2}{c |}{0.999} &
        \multicolumn{2}{c |}{0.999} \\
    }
    
    \par
    \medskip
    
    \ResultTable{Results for Inferable Relations}{ \rowcolor{\RowColor}
        0 &
        locatedIn &
            0.987 &
            0.991 &
            0.999 &
            0.986 &
            0.999 &
            800 / 10800 \\
        1 &
        neighborOf &
            --- &
            --- &
            0.999 &
            --- &
            0.999 &
            0 / 173072 \\
    }{
        \multicolumn{2}{c |}{0.929} &
        \multicolumn{2}{c |}{0.991} &
        \multicolumn{2}{c |}{0.999} \\
    }

\end{center}


\subsection{Countries S3}

\medskip

\begin{center}

    \ResultTable{Results for Inferable Classes}{ \rowcolor{\RowColor}
        0             &
        country       &
            1.000     &
            1.000     &
            1.000     &
            1.000     &
            1.000     &
            400 / 580 \\
        1             &
        region        &
            1.000     &
            1.000     &
            1.000     &
            1.000     &
            1.000     &
            120 / 860 \\ \rowcolor{\RowColor}
        2             &
        subregion     &
            1.000     &
            1.000     &
            1.000     &
            1.000     &
            1.000     &
            460 / 520 \\
    }{
        \multicolumn{2}{c |}{1.000} &
        \multicolumn{2}{c |}{1.000} &
        \multicolumn{2}{c |}{1.000} \\
    }
    
    \par
    \medskip
    
    \ResultTable{Results for Specified Relations}{ \rowcolor{\RowColor}
        0 &
        locatedIn &
            0.993 &
            0.999 &
            0.994 &
            0.994 &
            --- &
            7820 / 0 \\
        1 &
        neighborOf &
            1.000 &
            1.000 &
            1.000 &
            1.000 &
            --- &
            10216 / 0 \\
    }{
        \multicolumn{2}{c |}{0.996} &
        \multicolumn{2}{c |}{0.999} &
        \multicolumn{2}{c |}{0.999} \\
    }
    
    \par
    \medskip
    
    \ResultTable{Results for Inferable Relations}{ \rowcolor{\RowColor}
        0 &
        locatedIn &
            0.930 &
            0.988 &
            0.998 &
            0.981 &
            0.998 &
            800 / 10800 \\
        1 &
        neighborOf &
            --- &
            --- &
            0.999 &
            --- &
            0.999 &
            0 / 172966 \\
    }{
        \multicolumn{2}{c |}{0.916} &
        \multicolumn{2}{c |}{0.988} &
        \multicolumn{2}{c |}{0.999} \\
    }

\end{center}


\vskip 0.2in
\bibliography{literature}

\begin{thebibliography}{}

\bibitem [\protect \citeauthoryear {%
Bench-Capon%
\ \BBA {} Dunne%
}{%
Bench-Capon%
\ \BBA {} Dunne%
}{%
{\protect \APACyear {2007}}%
}]{%
bench2007argumentation}
\APACinsertmetastar {%
bench2007argumentation}%
\begin{APACrefauthors}%
Bench-Capon, T\BPBI J\BPBI M.%
\BCBT {}\ \BBA {} Dunne, P\BPBI E.%
\end{APACrefauthors}%
\unskip\
\newblock
\APACrefYearMonthDay{2007}{}{}.
\newblock
{\BBOQ}\APACrefatitle {Argumentation in artificial intelligence} {Argumentation
  in artificial intelligence}.{\BBCQ}
\newblock
\APACjournalVolNumPages{Artificial Intelligence}{171}{10-15}{619--641}.
\PrintBackRefs{\CurrentBib}

\bibitem [\protect \citeauthoryear {%
Bengio%
}{%
Bengio%
}{%
{\protect \APACyear {2009}}%
}]{%
Bengio2009}
\APACinsertmetastar {%
Bengio2009}%
\begin{APACrefauthors}%
Bengio, Y.%
\end{APACrefauthors}%
\unskip\
\newblock
\APACrefYearMonthDay{2009}{}{}.
\newblock
{\BBOQ}\APACrefatitle {Learning Deep Architectures for {AI}} {Learning deep
  architectures for {AI}}.{\BBCQ}
\newblock
\APACjournalVolNumPages{Foundations and Trends in Machine
  Learning}{2}{1}{1--127}.
\PrintBackRefs{\CurrentBib}

\bibitem [\protect \citeauthoryear {%
Bizer%
\ \protect \BOthers {.}}{%
Bizer%
\ \protect \BOthers {.}}{%
{\protect \APACyear {2009}}%
}]{%
Bizer2009}
\APACinsertmetastar {%
Bizer2009}%
\begin{APACrefauthors}%
Bizer, C.%
, Lehmann, J.%
, Kobilarov, G.%
, Auer, S.%
, Becker, C.%
, Cyganiak, R.%
\BCBL {}\ \BBA {} Hellmann, S.%
\end{APACrefauthors}%
\unskip\
\newblock
\APACrefYearMonthDay{2009}{}{}.
\newblock
{\BBOQ}\APACrefatitle {{DBpedia -- A} crystallization point for the {Web of
  Data}} {{DBpedia -- A} crystallization point for the {Web of Data}}.{\BBCQ}
\newblock
\APACjournalVolNumPages{Journal of Web Semantics}{7}{3}{154--165}.
\PrintBackRefs{\CurrentBib}

\bibitem [\protect \citeauthoryear {%
Bouchard%
, Singh%
\BCBL {}\ \BBA {} Trouillon%
}{%
Bouchard%
\ \protect \BOthers {.}}{%
{\protect \APACyear {2015}}%
}]{%
Bouchard2015}
\APACinsertmetastar {%
Bouchard2015}%
\begin{APACrefauthors}%
Bouchard, G.%
, Singh, S.%
\BCBL {}\ \BBA {} Trouillon, T.%
\end{APACrefauthors}%
\unskip\
\newblock
\APACrefYearMonthDay{2015}{}{}.
\newblock
{\BBOQ}\APACrefatitle {On approximate reasoning capabilities of low-rank vector
  spaces} {On approximate reasoning capabilities of low-rank vector
  spaces}.{\BBCQ}
\newblock
\BIn{} \APACrefbtitle {{Proceedings of the 2015 AAAI Spring Symposium on
  Knowledge Representation and Reasoning: Integrating Symbolic and Neural
  Approaches}} {{Proceedings of the 2015 AAAI Spring Symposium on Knowledge
  Representation and Reasoning: Integrating Symbolic and Neural Approaches}}\
  (\BPGS\ 6--9).
\PrintBackRefs{\CurrentBib}

\bibitem [\protect \citeauthoryear {%
Bowman%
, Potts%
\BCBL {}\ \BBA {} Manning%
}{%
Bowman%
\ \protect \BOthers {.}}{%
{\protect \APACyear {2014}}%
}]{%
Bowman2014}
\APACinsertmetastar {%
Bowman2014}%
\begin{APACrefauthors}%
Bowman, S\BPBI R.%
, Potts, C.%
\BCBL {}\ \BBA {} Manning, C\BPBI D.%
\end{APACrefauthors}%
\unskip\
\newblock
\APACrefYearMonthDay{2014}{}{}.
\newblock
{\BBOQ}\APACrefatitle {Recursive Neural Networks Can Learn Logical Semantics}
  {Recursive neural networks can learn logical semantics}.{\BBCQ}
\newblock
\APACjournalVolNumPages{arXiv preprint arXiv:1406.1827}{}{}{}.
\PrintBackRefs{\CurrentBib}

\bibitem [\protect \citeauthoryear {%
Bowman%
, Potts%
\BCBL {}\ \BBA {} Manning%
}{%
Bowman%
\ \protect \BOthers {.}}{%
{\protect \APACyear {2015}}%
}]{%
Bowman2015}
\APACinsertmetastar {%
Bowman2015}%
\begin{APACrefauthors}%
Bowman, S\BPBI R.%
, Potts, C.%
\BCBL {}\ \BBA {} Manning, C\BPBI D.%
\end{APACrefauthors}%
\unskip\
\newblock
\APACrefYearMonthDay{2015}{}{}.
\newblock
{\BBOQ}\APACrefatitle {{Learning distributed word representations for natural
  logic reasoning}} {{Learning distributed word representations for natural
  logic reasoning}}.{\BBCQ}
\newblock
\BIn{} \APACrefbtitle {{Proceedings of the 2015 AAAI Spring Symposium on
  Knowledge Representation and Reasoning: Integrating Symbolic and Neural
  Approaches}} {{Proceedings of the 2015 AAAI Spring Symposium on Knowledge
  Representation and Reasoning: Integrating Symbolic and Neural Approaches}}\
  (\BPGS\ 10--13).
\PrintBackRefs{\CurrentBib}

\bibitem [\protect \citeauthoryear {%
Cai%
, Ke%
, Xu%
\BCBL {}\ \BBA {} Su%
}{%
Cai%
\ \protect \BOthers {.}}{%
{\protect \APACyear {2017}}%
}]{%
Cai2017}
\APACinsertmetastar {%
Cai2017}%
\begin{APACrefauthors}%
Cai, C\BHBI H.%
, Ke, D.%
, Xu, Y.%
\BCBL {}\ \BBA {} Su, K.%
\end{APACrefauthors}%
\unskip\
\newblock
\APACrefYearMonthDay{2017}{}{}.
\newblock
{\BBOQ}\APACrefatitle {Symbolic Manipulation Based on Deep Neural Networks and
  its Application to Axiom Discovery} {Symbolic manipulation based on deep
  neural networks and its application to axiom discovery}.{\BBCQ}
\newblock
\BIn{} \APACrefbtitle {{Proceedings of the 2017 International Joint Conference
  on Neural Networks}} {{Proceedings of the 2017 International Joint Conference
  on Neural Networks}}\ (\BPGS\ 2136--2143).
\PrintBackRefs{\CurrentBib}

\bibitem [\protect \citeauthoryear {%
Cal{\`{\i}}%
, Gottlob%
\BCBL {}\ \BBA {} Lukasiewicz%
}{%
Cal{\`{\i}}%
\ \protect \BOthers {.}}{%
{\protect \APACyear {2012}}%
}]{%
DBLP:journals/ws/CaliGL12}
\APACinsertmetastar {%
DBLP:journals/ws/CaliGL12}%
\begin{APACrefauthors}%
Cal{\`{\i}}, A.%
, Gottlob, G.%
\BCBL {}\ \BBA {} Lukasiewicz, T.%
\end{APACrefauthors}%
\unskip\
\newblock
\APACrefYearMonthDay{2012}{}{}.
\newblock
{\BBOQ}\APACrefatitle {A general {D}atalog-based framework for tractable query
  answering over ontologies} {A general {D}atalog-based framework for tractable
  query answering over ontologies}.{\BBCQ}
\newblock
\APACjournalVolNumPages{Journal of Web Semantics}{14}{}{57--83}.
\PrintBackRefs{\CurrentBib}

\bibitem [\protect \citeauthoryear {%
Cingillioglu%
\ \BBA {} Russo%
}{%
Cingillioglu%
\ \BBA {} Russo%
}{%
{\protect \APACyear {2018}}%
}]{%
Cingillioglu2018}
\APACinsertmetastar {%
Cingillioglu2018}%
\begin{APACrefauthors}%
Cingillioglu, N.%
\BCBT {}\ \BBA {} Russo, A.%
\end{APACrefauthors}%
\unskip\
\newblock
\APACrefYearMonthDay{2018}{}{}.
\newblock
{\BBOQ}\APACrefatitle {{DeepLogic: E}nd-to-End Logical Reasoning} {{DeepLogic:
  E}nd-to-end logical reasoning}.{\BBCQ}
\newblock
\APACjournalVolNumPages{arXiv preprint arXiv:1805.07433}{}{}{}.
\PrintBackRefs{\CurrentBib}

\bibitem [\protect \citeauthoryear {%
Cohen%
}{%
Cohen%
}{%
{\protect \APACyear {2016}}%
}]{%
Cohen2016}
\APACinsertmetastar {%
Cohen2016}%
\begin{APACrefauthors}%
Cohen, W\BPBI W.%
\end{APACrefauthors}%
\unskip\
\newblock
\APACrefYearMonthDay{2016}{}{}.
\newblock
{\BBOQ}\APACrefatitle {{TensorLog: A} Differentiable Deductive Database}
  {{TensorLog: A} differentiable deductive database}.{\BBCQ}
\newblock
\APACjournalVolNumPages{arXiv preprint arXiv:1605.06523}{}{}{}.
\PrintBackRefs{\CurrentBib}

\bibitem [\protect \citeauthoryear {%
Cohen%
, Yang%
\BCBL {}\ \BBA {} Rivard~Mazaitis%
}{%
Cohen%
\ \protect \BOthers {.}}{%
{\protect \APACyear {2017}}%
}]{%
Cohen2017}
\APACinsertmetastar {%
Cohen2017}%
\begin{APACrefauthors}%
Cohen, W\BPBI W.%
, Yang, F.%
\BCBL {}\ \BBA {} Rivard~Mazaitis, K.%
\end{APACrefauthors}%
\unskip\
\newblock
\APACrefYearMonthDay{2017}{}{}.
\newblock
{\BBOQ}\APACrefatitle {{TensorLog: D}eep Learning Meets Probabilistic {DBs}}
  {{TensorLog: D}eep learning meets probabilistic {DBs}}.{\BBCQ}
\newblock
\APACjournalVolNumPages{arXiv preprint arXiv:1707.05390}{}{}{}.
\PrintBackRefs{\CurrentBib}

\bibitem [\protect \citeauthoryear {%
Dai%
, Xu%
, Yu%
\BCBL {}\ \BBA {} Zhou%
}{%
Dai%
\ \protect \BOthers {.}}{%
{\protect \APACyear {2018}}%
}]{%
Dai2018}
\APACinsertmetastar {%
Dai2018}%
\begin{APACrefauthors}%
Dai, W\BHBI Z.%
, Xu, Q\BHBI L.%
, Yu, Y.%
\BCBL {}\ \BBA {} Zhou, Z\BHBI H.%
\end{APACrefauthors}%
\unskip\
\newblock
\APACrefYearMonthDay{2018}{}{}.
\newblock
{\BBOQ}\APACrefatitle {Tunneling Neural Perception and Logic Reasoning through
  Abductive Learning} {Tunneling neural perception and logic reasoning through
  abductive learning}.{\BBCQ}
\newblock
\APACjournalVolNumPages{arXiv preprint arXiv:1802.01173}{}{}{}.
\PrintBackRefs{\CurrentBib}

\bibitem [\protect \citeauthoryear {%
d'Avila Garcez%
\ \protect \BOthers {.}}{%
d'Avila Garcez%
\ \protect \BOthers {.}}{%
{\protect \APACyear {2015}}%
}]{%
dAvila2015}
\APACinsertmetastar {%
dAvila2015}%
\begin{APACrefauthors}%
d'Avila Garcez, A\BPBI S.%
, Besold, T\BPBI R.%
, De~Raedt, L.%
, F{\"o}ldiak, P.%
, Hitzler, P.%
, Icard, T.%
\BDBL {}Silver, D\BPBI L.%
\end{APACrefauthors}%
\unskip\
\newblock
\APACrefYearMonthDay{2015}{}{}.
\newblock
{\BBOQ}\APACrefatitle {Neural-Symbolic Learning and Reasoning: {C}ontributions
  and Challenges} {Neural-symbolic learning and reasoning: {C}ontributions and
  challenges}.{\BBCQ}
\newblock
\BIn{} \APACrefbtitle {{Proceedings of the 2015 AAAI Spring Symposium on
  Knowledge Representation and Reasoning: Integrating Symbolic and Neural
  Approaches}} {{Proceedings of the 2015 AAAI Spring Symposium on Knowledge
  Representation and Reasoning: Integrating Symbolic and Neural Approaches}}\
  (\BPGS\ 18--21).
\PrintBackRefs{\CurrentBib}

\bibitem [\protect \citeauthoryear {%
d'Avila Garcez%
, Broda%
\BCBL {}\ \BBA {} Gabbay%
}{%
d'Avila Garcez%
\ \protect \BOthers {.}}{%
{\protect \APACyear {2012}}%
}]{%
dAvila2012}
\APACinsertmetastar {%
dAvila2012}%
\begin{APACrefauthors}%
d'Avila Garcez, A\BPBI S.%
, Broda, K\BPBI B.%
\BCBL {}\ \BBA {} Gabbay, D\BPBI M.%
\end{APACrefauthors}%
\unskip\
\newblock
\APACrefYear{2012}.
\newblock
\APACrefbtitle {{Neural-Symbolic Learning Systems: Foundations and
  Applications}} {{Neural-Symbolic Learning Systems: Foundations and
  Applications}}.
\newblock
\APACaddressPublisher{}{Springer Science \& Business Media}.
\PrintBackRefs{\CurrentBib}

\bibitem [\protect \citeauthoryear {%
d'Avila Garcez%
\ \BBA {} Zaverucha%
}{%
d'Avila Garcez%
\ \BBA {} Zaverucha%
}{%
{\protect \APACyear {1999}}%
}]{%
dAvila1999}
\APACinsertmetastar {%
dAvila1999}%
\begin{APACrefauthors}%
d'Avila Garcez, A\BPBI S.%
\BCBT {}\ \BBA {} Zaverucha, G.%
\end{APACrefauthors}%
\unskip\
\newblock
\APACrefYearMonthDay{1999}{}{}.
\newblock
{\BBOQ}\APACrefatitle {The Connectionist Inductive Learning and Logic
  Programming System} {The connectionist inductive learning and logic
  programming system}.{\BBCQ}
\newblock
\APACjournalVolNumPages{Applied Intelligence}{11}{}{59--77}.
\PrintBackRefs{\CurrentBib}

\bibitem [\protect \citeauthoryear {%
Demeester%
, Rockt{\"a}schel%
\BCBL {}\ \BBA {} Riedel%
}{%
Demeester%
\ \protect \BOthers {.}}{%
{\protect \APACyear {2016}}%
}]{%
Demeester2016}
\APACinsertmetastar {%
Demeester2016}%
\begin{APACrefauthors}%
Demeester, T.%
, Rockt{\"a}schel, T.%
\BCBL {}\ \BBA {} Riedel, S.%
\end{APACrefauthors}%
\unskip\
\newblock
\APACrefYearMonthDay{2016}{}{}.
\newblock
{\BBOQ}\APACrefatitle {Lifted Rule Injection for Relation Embeddings} {Lifted
  rule injection for relation embeddings}.{\BBCQ}
\newblock
\BIn{} \APACrefbtitle {{Proceedings of the 2016 Conference on Empirical Methods
  in Natural Language Processing}} {{Proceedings of the 2016 Conference on
  Empirical Methods in Natural Language Processing}}\ (\BPGS\ 1389--1399).
\PrintBackRefs{\CurrentBib}

\bibitem [\protect \citeauthoryear {%
Diligenti%
, Gori%
, Maggini%
\BCBL {}\ \BBA {} Rigutini%
}{%
Diligenti%
\ \protect \BOthers {.}}{%
{\protect \APACyear {2012}}%
}]{%
Diligenti2012}
\APACinsertmetastar {%
Diligenti2012}%
\begin{APACrefauthors}%
Diligenti, M.%
, Gori, M.%
, Maggini, M.%
\BCBL {}\ \BBA {} Rigutini, L.%
\end{APACrefauthors}%
\unskip\
\newblock
\APACrefYearMonthDay{2012}{}{}.
\newblock
{\BBOQ}\APACrefatitle {{Bridging logic and kernel machines}} {{Bridging logic
  and kernel machines}}.{\BBCQ}
\newblock
\APACjournalVolNumPages{Machine Learning}{86}{}{57--88}.
\PrintBackRefs{\CurrentBib}

\bibitem [\protect \citeauthoryear {%
Ding%
}{%
Ding%
}{%
{\protect \APACyear {1995}}%
}]{%
Ding1995}
\APACinsertmetastar {%
Ding1995}%
\begin{APACrefauthors}%
Ding, L.%
\end{APACrefauthors}%
\unskip\
\newblock
\APACrefYearMonthDay{1995}{}{}.
\newblock
{\BBOQ}\APACrefatitle {{Neural Prolog---The concepts, construction and
  mechanism}} {{Neural Prolog---The concepts, construction and
  mechanism}}.{\BBCQ}
\newblock
\BIn{} \APACrefbtitle {{Proceedings of the 1995 IEEE International Conference
  on Systems, Man and Cybernetics. Intelligent Systems for the 21st Century}}
  {{Proceedings of the 1995 IEEE International Conference on Systems, Man and
  Cybernetics. Intelligent Systems for the 21st Century}}\ (\BVOL~4, \BPGS\
  3603--3608).
\PrintBackRefs{\CurrentBib}

\bibitem [\protect \citeauthoryear {%
Donadello%
, Serafini%
\BCBL {}\ \BBA {} d'Avila Garcez%
}{%
Donadello%
\ \protect \BOthers {.}}{%
{\protect \APACyear {2017}}%
}]{%
Donadello2017}
\APACinsertmetastar {%
Donadello2017}%
\begin{APACrefauthors}%
Donadello, I.%
, Serafini, L.%
\BCBL {}\ \BBA {} d'Avila Garcez, A.%
\end{APACrefauthors}%
\unskip\
\newblock
\APACrefYearMonthDay{2017}{}{}.
\newblock
{\BBOQ}\APACrefatitle {Logic Tensor Networks for Semantic Image Interpretation}
  {Logic tensor networks for semantic image interpretation}.{\BBCQ}
\newblock
\BIn{} \APACrefbtitle {{Proceedings of the 26th International Joint Conference
  on Artificial Intelligence}} {{Proceedings of the 26th International Joint
  Conference on Artificial Intelligence}}\ (\BPGS\ 1596--1602).
\PrintBackRefs{\CurrentBib}

\bibitem [\protect \citeauthoryear {%
Dong%
\ \protect \BOthers {.}}{%
Dong%
\ \protect \BOthers {.}}{%
{\protect \APACyear {2014}}%
}]{%
Dong2014}
\APACinsertmetastar {%
Dong2014}%
\begin{APACrefauthors}%
Dong, X\BPBI L.%
, Gabrilovich, E.%
, Heitz, G.%
, Horn, W.%
, Lao, N.%
, Murphy, K.%
\BDBL {}Zhang, W.%
\end{APACrefauthors}%
\unskip\
\newblock
\APACrefYearMonthDay{2014}{}{}.
\newblock
{\BBOQ}\APACrefatitle {{Knowledge Vault: A W}eb-Scale Approach to Probabilistic
  Knowledge Fusion} {{Knowledge Vault: A W}eb-scale approach to probabilistic
  knowledge fusion}.{\BBCQ}
\newblock
\BIn{} \APACrefbtitle {{Proceedings of the 20th ACM SIGKDD International
  Conference on Knowledge Discovery and Data Mining}} {{Proceedings of the 20th
  ACM SIGKDD International Conference on Knowledge Discovery and Data Mining}}\
  (\BPGS\ 601--610).
\PrintBackRefs{\CurrentBib}

\bibitem [\protect \citeauthoryear {%
Ebrahimi%
\ \protect \BOthers {.}}{%
Ebrahimi%
\ \protect \BOthers {.}}{%
{\protect \APACyear {2018}}%
}]{%
Ebrahimi2018}
\APACinsertmetastar {%
Ebrahimi2018}%
\begin{APACrefauthors}%
Ebrahimi, M.%
, Sarker, M\BPBI K.%
, Bianchi, F.%
, Xie, N.%
, Doran, D.%
\BCBL {}\ \BBA {} Hitzler, P.%
\end{APACrefauthors}%
\unskip\
\newblock
\APACrefYearMonthDay{2018}{}{}.
\newblock
{\BBOQ}\APACrefatitle {Reasoning over {RDF} Knowledge Bases using Deep
  Learning} {Reasoning over {RDF} knowledge bases using deep learning}.{\BBCQ}
\newblock
\APACjournalVolNumPages{arXiv preprint arXiv:1811.04132}{}{}{}.
\PrintBackRefs{\CurrentBib}

\bibitem [\protect \citeauthoryear {%
Eiter%
\ \BBA {} Lukasiewicz%
}{%
Eiter%
\ \BBA {} Lukasiewicz%
}{%
{\protect \APACyear {2000}}%
}]{%
eiter2000default}
\APACinsertmetastar {%
eiter2000default}%
\begin{APACrefauthors}%
Eiter, T.%
\BCBT {}\ \BBA {} Lukasiewicz, T.%
\end{APACrefauthors}%
\unskip\
\newblock
\APACrefYearMonthDay{2000}{}{}.
\newblock
{\BBOQ}\APACrefatitle {Default reasoning from conditional knowledge bases:
  Complexity and tractable cases} {Default reasoning from conditional knowledge
  bases: Complexity and tractable cases}.{\BBCQ}
\newblock
\APACjournalVolNumPages{Artificial Intelligence}{124}{2}{169--241}.
\PrintBackRefs{\CurrentBib}

\bibitem [\protect \citeauthoryear {%
Evans%
, Saxton%
, Amos%
, Kohli%
\BCBL {}\ \BBA {} Grefenstette%
}{%
Evans%
\ \protect \BOthers {.}}{%
{\protect \APACyear {2018}}%
}]{%
Evans2018}
\APACinsertmetastar {%
Evans2018}%
\begin{APACrefauthors}%
Evans, R.%
, Saxton, D.%
, Amos, D.%
, Kohli, P.%
\BCBL {}\ \BBA {} Grefenstette, E.%
\end{APACrefauthors}%
\unskip\
\newblock
\APACrefYearMonthDay{2018}{}{}.
\newblock
{\BBOQ}\APACrefatitle {Can Neural Networks Understand Logical Entailment?} {Can
  neural networks understand logical entailment?}{\BBCQ}
\newblock
\BIn{} \APACrefbtitle {{Proceedings of the 6th International Conference on
  Learning Representations}.} {{Proceedings of the 6th International Conference
  on Learning Representations}.}
\PrintBackRefs{\CurrentBib}

\bibitem [\protect \citeauthoryear {%
Fran{\c{c}}a%
, Zaverucha%
\BCBL {}\ \BBA {} d'Avila Garcez%
}{%
Fran{\c{c}}a%
\ \protect \BOthers {.}}{%
{\protect \APACyear {2014}}%
}]{%
Franca2014}
\APACinsertmetastar {%
Franca2014}%
\begin{APACrefauthors}%
Fran{\c{c}}a, M\BPBI V\BPBI M.%
, Zaverucha, G.%
\BCBL {}\ \BBA {} d'Avila Garcez, A\BPBI S.%
\end{APACrefauthors}%
\unskip\
\newblock
\APACrefYearMonthDay{2014}{}{}.
\newblock
{\BBOQ}\APACrefatitle {{Fast relational learning using bottom clause
  propositionalization with artificial neural networks}} {{Fast relational
  learning using bottom clause propositionalization with artificial neural
  networks}}.{\BBCQ}
\newblock
\APACjournalVolNumPages{Machine Learning}{94}{1}{81--104}.
\PrintBackRefs{\CurrentBib}

\bibitem [\protect \citeauthoryear {%
Goodfellow%
\ \protect \BOthers {.}}{%
Goodfellow%
\ \protect \BOthers {.}}{%
{\protect \APACyear {2014}}%
}]{%
Goodfellow2014}
\APACinsertmetastar {%
Goodfellow2014}%
\begin{APACrefauthors}%
Goodfellow, I\BPBI J.%
, Pouget-Abadie, J.%
, Mirza, M.%
, Xu, B.%
, Warde-Farley, D.%
, Ozair, S.%
\BDBL {}Bengio, Y.%
\end{APACrefauthors}%
\unskip\
\newblock
\APACrefYearMonthDay{2014}{}{}.
\newblock
{\BBOQ}\APACrefatitle {Generative Adversarial Nets} {Generative adversarial
  nets}.{\BBCQ}
\newblock
\BIn{} \APACrefbtitle {{Advances in Neural Information Processing Systems 27}.}
  {{Advances in Neural Information Processing Systems 27}.}
\PrintBackRefs{\CurrentBib}

\bibitem [\protect \citeauthoryear {%
Graves%
\ \protect \BOthers {.}}{%
Graves%
\ \protect \BOthers {.}}{%
{\protect \APACyear {2016}}%
}]{%
Graves2016}
\APACinsertmetastar {%
Graves2016}%
\begin{APACrefauthors}%
Graves, A.%
, Wayne, G.%
, Reynolds, M.%
, Harley, T.%
, Danihelka, I.%
, Grabska-Barwi{\'{n}}ska, A.%
\BDBL {}Hassabis, D.%
\end{APACrefauthors}%
\unskip\
\newblock
\APACrefYearMonthDay{2016}{}{}.
\newblock
{\BBOQ}\APACrefatitle {{Hybrid computing using a neural network with dynamic
  external memory}} {{Hybrid computing using a neural network with dynamic
  external memory}}.{\BBCQ}
\newblock
\APACjournalVolNumPages{Nature}{538}{}{471--476}.
\PrintBackRefs{\CurrentBib}

\bibitem [\protect \citeauthoryear {%
Guo%
, Pan%
\BCBL {}\ \BBA {} Heflin%
}{%
Guo%
\ \protect \BOthers {.}}{%
{\protect \APACyear {2005}}%
}]{%
Guo2005}
\APACinsertmetastar {%
Guo2005}%
\begin{APACrefauthors}%
Guo, Y.%
, Pan, Z.%
\BCBL {}\ \BBA {} Heflin, J.%
\end{APACrefauthors}%
\unskip\
\newblock
\APACrefYearMonthDay{2005}{}{}.
\newblock
{\BBOQ}\APACrefatitle {{LUBM: A benchmark for OWL knowledge base systems}}
  {{LUBM: A benchmark for OWL knowledge base systems}}.{\BBCQ}
\newblock
\APACjournalVolNumPages{Journal of Web Semantics}{3}{2/3}{158--182}.
\PrintBackRefs{\CurrentBib}

\bibitem [\protect \citeauthoryear {%
Hammer%
\ \BBA {} Hitzler%
}{%
Hammer%
\ \BBA {} Hitzler%
}{%
{\protect \APACyear {2007}}%
}]{%
Hammer2007}
\APACinsertmetastar {%
Hammer2007}%
\begin{APACrefauthors}%
Hammer, B.%
\BCBT {}\ \BBA {} Hitzler, P.%
\end{APACrefauthors}%
\unskip\
\newblock
\APACrefYear{2007}.
\newblock
\APACrefbtitle {{Perspectives of Neural-Symbolic Integration}} {{Perspectives
  of Neural-Symbolic Integration}}.
\newblock
\APACaddressPublisher{}{Springer}.
\PrintBackRefs{\CurrentBib}

\bibitem [\protect \citeauthoryear {%
Hassabis%
, Kumaran%
, Summerfield%
\BCBL {}\ \BBA {} Botvinick%
}{%
Hassabis%
\ \protect \BOthers {.}}{%
{\protect \APACyear {2017}}%
}]{%
Hassabis2017}
\APACinsertmetastar {%
Hassabis2017}%
\begin{APACrefauthors}%
Hassabis, D.%
, Kumaran, D.%
, Summerfield, C.%
\BCBL {}\ \BBA {} Botvinick, M.%
\end{APACrefauthors}%
\unskip\
\newblock
\APACrefYearMonthDay{2017}{}{}.
\newblock
{\BBOQ}\APACrefatitle {Neuroscience-inspired artificial intelligence}
  {Neuroscience-inspired artificial intelligence}.{\BBCQ}
\newblock
\APACjournalVolNumPages{Neuron}{95}{2}{245--258}.
\PrintBackRefs{\CurrentBib}

\bibitem [\protect \citeauthoryear {%
Henaff%
, Weston%
, Szlam%
, Bordes%
\BCBL {}\ \BBA {} LeCun%
}{%
Henaff%
\ \protect \BOthers {.}}{%
{\protect \APACyear {2017}}%
}]{%
Henaff2017}
\APACinsertmetastar {%
Henaff2017}%
\begin{APACrefauthors}%
Henaff, M.%
, Weston, J.%
, Szlam, A.%
, Bordes, A.%
\BCBL {}\ \BBA {} LeCun, Y.%
\end{APACrefauthors}%
\unskip\
\newblock
\APACrefYearMonthDay{2017}{}{}.
\newblock
{\BBOQ}\APACrefatitle {Tracking the World State with Recurrent Entity Networks}
  {Tracking the world state with recurrent entity networks}.{\BBCQ}
\newblock
\APACjournalVolNumPages{arXiv preprint arXiv:1612.03969}{}{}{}.
\PrintBackRefs{\CurrentBib}

\bibitem [\protect \citeauthoryear {%
Hu%
, Ma%
, Liu%
, Hovy%
\BCBL {}\ \BBA {} Xing%
}{%
Hu%
\ \protect \BOthers {.}}{%
{\protect \APACyear {2016}}%
}]{%
Hu2016}
\APACinsertmetastar {%
Hu2016}%
\begin{APACrefauthors}%
Hu, Z.%
, Ma, X.%
, Liu, Z.%
, Hovy, E.%
\BCBL {}\ \BBA {} Xing, E.%
\end{APACrefauthors}%
\unskip\
\newblock
\APACrefYearMonthDay{2016}{}{}.
\newblock
{\BBOQ}\APACrefatitle {Harnessing Deep Neural Networks with Logic Rules}
  {Harnessing deep neural networks with logic rules}.{\BBCQ}
\newblock
\BIn{} \APACrefbtitle {{Proceedings of the 54th Annual Meeting of the
  Association for Computational Linguistics (Volume 1: Long Papers)}}
  {{Proceedings of the 54th Annual Meeting of the Association for Computational
  Linguistics (Volume 1: Long Papers)}}\ (\BPGS\ 2410--2420).
\PrintBackRefs{\CurrentBib}

\bibitem [\protect \citeauthoryear {%
Kingma%
\ \BBA {} Ba%
}{%
Kingma%
\ \BBA {} Ba%
}{%
{\protect \APACyear {2015}}%
}]{%
Kingma2015}
\APACinsertmetastar {%
Kingma2015}%
\begin{APACrefauthors}%
Kingma, D\BPBI P.%
\BCBT {}\ \BBA {} Ba, J\BPBI L.%
\end{APACrefauthors}%
\unskip\
\newblock
\APACrefYearMonthDay{2015}{}{}.
\newblock
{\BBOQ}\APACrefatitle {{Adam: A} Method for Stochastic Optimization} {{Adam: A}
  method for stochastic optimization}.{\BBCQ}
\newblock
\BIn{} \APACrefbtitle {{Proceedings of the 3rd International Conference on
  Learning Representations}.} {{Proceedings of the 3rd International Conference
  on Learning Representations}.}
\PrintBackRefs{\CurrentBib}

\bibitem [\protect \citeauthoryear {%
Kok%
\ \BBA {} Domingos%
}{%
Kok%
\ \BBA {} Domingos%
}{%
{\protect \APACyear {2007}}%
}]{%
Kok2007}
\APACinsertmetastar {%
Kok2007}%
\begin{APACrefauthors}%
Kok, S.%
\BCBT {}\ \BBA {} Domingos, P.%
\end{APACrefauthors}%
\unskip\
\newblock
\APACrefYearMonthDay{2007}{}{}.
\newblock
{\BBOQ}\APACrefatitle {{Statistical predicate invention}} {{Statistical
  predicate invention}}.{\BBCQ}
\newblock
\BIn{} \APACrefbtitle {{Proceedings of the 24th International Conference on
  Machine Learning}} {{Proceedings of the 24th International Conference on
  Machine Learning}}\ (\BPGS\ 433--440).
\PrintBackRefs{\CurrentBib}

\bibitem [\protect \citeauthoryear {%
Lembo%
, Lenzerini%
, Rosati%
, Ruzzi%
\BCBL {}\ \BBA {} Savo%
}{%
Lembo%
\ \protect \BOthers {.}}{%
{\protect \APACyear {2010}}%
}]{%
LLRRS10}
\APACinsertmetastar {%
LLRRS10}%
\begin{APACrefauthors}%
Lembo, D.%
, Lenzerini, M.%
, Rosati, R.%
, Ruzzi, M.%
\BCBL {}\ \BBA {} Savo, D\BPBI F.%
\end{APACrefauthors}%
\unskip\
\newblock
\APACrefYearMonthDay{2010}{}{}.
\newblock
{\BBOQ}\APACrefatitle {Inconsistency-Tolerant Semantics for Description Logics}
  {Inconsistency-tolerant semantics for description logics}.{\BBCQ}
\newblock
\BIn{} \APACrefbtitle {{Proceedings of the 4th International Conference on Web
  Reasoning and Rule Systems}} {{Proceedings of the 4th International
  Conference on Web Reasoning and Rule Systems}}\ (\BPGS\ 103--117).
\PrintBackRefs{\CurrentBib}

\bibitem [\protect \citeauthoryear {%
Lukasiewicz%
, Martinez%
, Pieris%
\BCBL {}\ \BBA {} Simari%
}{%
Lukasiewicz%
\ \protect \BOthers {.}}{%
{\protect \APACyear {2015}}%
}]{%
LMPS15}
\APACinsertmetastar {%
LMPS15}%
\begin{APACrefauthors}%
Lukasiewicz, T.%
, Martinez, M\BPBI V.%
, Pieris, A.%
\BCBL {}\ \BBA {} Simari, G\BPBI I.%
\end{APACrefauthors}%
\unskip\
\newblock
\APACrefYearMonthDay{2015}{}{}.
\newblock
{\BBOQ}\APACrefatitle {From Classical to Consistent Query Answering under
  Existential Rules} {From classical to consistent query answering under
  existential rules}.{\BBCQ}
\newblock
\BIn{} \APACrefbtitle {{Proceedings of the 29th AAAI Conference on Artificial
  Intelligence}} {{Proceedings of the 29th AAAI Conference on Artificial
  Intelligence}}\ (\BPGS\ 1546--1552).
\PrintBackRefs{\CurrentBib}

\bibitem [\protect \citeauthoryear {%
Makni%
\ \BBA {} Hendler%
}{%
Makni%
\ \BBA {} Hendler%
}{%
{\protect \APACyear {2018}}%
}]{%
Makni2018}
\APACinsertmetastar {%
Makni2018}%
\begin{APACrefauthors}%
Makni, B.%
\BCBT {}\ \BBA {} Hendler, J.%
\end{APACrefauthors}%
\unskip\
\newblock
\APACrefYearMonthDay{2018}{}{}.
\newblock
{\BBOQ}\APACrefatitle {{Deep learning for noise-tolerant RDFS reasoning}}
  {{Deep learning for noise-tolerant RDFS reasoning}}.{\BBCQ}
\newblock
\APACjournalVolNumPages{Semantic Web}{10}{5}{823--862}.
\PrintBackRefs{\CurrentBib}

\bibitem [\protect \citeauthoryear {%
Manhaeve%
, Duman{\v{c}}i{\'{c}}%
, Kimmig%
, Demeester%
\BCBL {}\ \BBA {} De~Raedt%
}{%
Manhaeve%
\ \protect \BOthers {.}}{%
{\protect \APACyear {2018}}%
}]{%
Manhaeve2018}
\APACinsertmetastar {%
Manhaeve2018}%
\begin{APACrefauthors}%
Manhaeve, R.%
, Duman{\v{c}}i{\'{c}}, S.%
, Kimmig, A.%
, Demeester, T.%
\BCBL {}\ \BBA {} De~Raedt, L.%
\end{APACrefauthors}%
\unskip\
\newblock
\APACrefYearMonthDay{2018}{}{}.
\newblock
{\BBOQ}\APACrefatitle {{DeepProbLog: N}eural Probabilistic Logic Programming}
  {{DeepProbLog: N}eural probabilistic logic programming}.{\BBCQ}
\newblock
\APACjournalVolNumPages{arXiv preprint arXiv:1805.10872}{}{}{}.
\PrintBackRefs{\CurrentBib}

\bibitem [\protect \citeauthoryear {%
Martin%
\ \BBA {} Chao%
}{%
Martin%
\ \BBA {} Chao%
}{%
{\protect \APACyear {2001}}%
}]{%
Martin2001}
\APACinsertmetastar {%
Martin2001}%
\begin{APACrefauthors}%
Martin, A.%
\BCBT {}\ \BBA {} Chao, L\BPBI L.%
\end{APACrefauthors}%
\unskip\
\newblock
\APACrefYearMonthDay{2001}{}{}.
\newblock
{\BBOQ}\APACrefatitle {Semantic memory and the brain: {S}tructure and
  processes} {Semantic memory and the brain: {S}tructure and processes}.{\BBCQ}
\newblock
\APACjournalVolNumPages{Current Opinion in Neurobiology}{11}{2}{194--201}.
\PrintBackRefs{\CurrentBib}

\bibitem [\protect \citeauthoryear {%
McCray%
}{%
McCray%
}{%
{\protect \APACyear {2003}}%
}]{%
McCray2003}
\APACinsertmetastar {%
McCray2003}%
\begin{APACrefauthors}%
McCray, A\BPBI T.%
\end{APACrefauthors}%
\unskip\
\newblock
\APACrefYearMonthDay{2003}{}{}.
\newblock
{\BBOQ}\APACrefatitle {An upper-level ontology for the biomedical domain} {An
  upper-level ontology for the biomedical domain}.{\BBCQ}
\newblock
\APACjournalVolNumPages{Comparative and Functional Genomics}{4}{}{80--84}.
\PrintBackRefs{\CurrentBib}

\bibitem [\protect \citeauthoryear {%
Middelburg%
}{%
Middelburg%
}{%
{\protect \APACyear {2011}}%
}]{%
Middelburg2011}
\APACinsertmetastar {%
Middelburg2011}%
\begin{APACrefauthors}%
Middelburg, C\BPBI A.%
\end{APACrefauthors}%
\unskip\
\newblock
\APACrefYearMonthDay{2011}{}{}.
\newblock
{\BBOQ}\APACrefatitle {A Survey of Paraconsistent Logics} {A survey of
  paraconsistent logics}.{\BBCQ}
\newblock
\APACjournalVolNumPages{arXiv preprint arXiv:1103.4324}{}{}{}.
\PrintBackRefs{\CurrentBib}

\bibitem [\protect \citeauthoryear {%
Mikolov%
, Corrado%
, Chen%
\BCBL {}\ \BBA {} Dean%
}{%
Mikolov%
\ \protect \BOthers {.}}{%
{\protect \APACyear {2013}}%
}]{%
Mikolov2013}
\APACinsertmetastar {%
Mikolov2013}%
\begin{APACrefauthors}%
Mikolov, T.%
, Corrado, G.%
, Chen, K.%
\BCBL {}\ \BBA {} Dean, J.%
\end{APACrefauthors}%
\unskip\
\newblock
\APACrefYearMonthDay{2013}{}{}.
\newblock
{\BBOQ}\APACrefatitle {Efficient Estimation of Word Representations in Vector
  Space} {Efficient estimation of word representations in vector space}.{\BBCQ}
\newblock
\BIn{} \APACrefbtitle {{Proceedings of the 1st International Conference on
  Learning Representations}.} {{Proceedings of the 1st International Conference
  on Learning Representations}.}
\PrintBackRefs{\CurrentBib}

\bibitem [\protect \citeauthoryear {%
Minervini%
, Bosnjak%
, Rockt{\"{a}}schel%
\BCBL {}\ \BBA {} Riedel%
}{%
Minervini%
\ \protect \BOthers {.}}{%
{\protect \APACyear {2018}}%
}]{%
Minervini2018}
\APACinsertmetastar {%
Minervini2018}%
\begin{APACrefauthors}%
Minervini, P.%
, Bosnjak, M.%
, Rockt{\"{a}}schel, T.%
\BCBL {}\ \BBA {} Riedel, S.%
\end{APACrefauthors}%
\unskip\
\newblock
\APACrefYearMonthDay{2018}{}{}.
\newblock
{\BBOQ}\APACrefatitle {Towards Neural Theorem Proving at Scale} {Towards neural
  theorem proving at scale}.{\BBCQ}
\newblock
\APACjournalVolNumPages{arXiv preprint arXiv:1807.08204}{}{}{}.
\PrintBackRefs{\CurrentBib}

\bibitem [\protect \citeauthoryear {%
Minervini%
, Costabello%
, Mu{\~{n}}oz%
, Nov{\'{a}}{\v{c}}ek%
\BCBL {}\ \BBA {} Vandebussche%
}{%
Minervini%
, Costabello%
\BCBL {}\ \protect \BOthers {.}}{%
{\protect \APACyear {2017}}%
}]{%
Minervini2017}
\APACinsertmetastar {%
Minervini2017}%
\begin{APACrefauthors}%
Minervini, P.%
, Costabello, L.%
, Mu{\~{n}}oz, E.%
, Nov{\'{a}}{\v{c}}ek, V.%
\BCBL {}\ \BBA {} Vandebussche, P\BHBI Y.%
\end{APACrefauthors}%
\unskip\
\newblock
\APACrefYearMonthDay{2017}{}{}.
\newblock
{\BBOQ}\APACrefatitle {Regularizing Knowledge Graph Embeddings via Equivalence
  and Inversion Axioms} {Regularizing knowledge graph embeddings via
  equivalence and inversion axioms}.{\BBCQ}
\newblock
\BIn{} \APACrefbtitle {{Machine Learning and Knowledge Discovery in Databases}}
  {{Machine Learning and Knowledge Discovery in Databases}}\ (\BPGS\ 668--683).
\PrintBackRefs{\CurrentBib}

\bibitem [\protect \citeauthoryear {%
Minervini%
, Demeester%
, Rockt{\"{a}}schel%
\BCBL {}\ \BBA {} Riedel%
}{%
Minervini%
, Demeester%
\BCBL {}\ \protect \BOthers {.}}{%
{\protect \APACyear {2017}}%
}]{%
Minervini2017a}
\APACinsertmetastar {%
Minervini2017a}%
\begin{APACrefauthors}%
Minervini, P.%
, Demeester, T.%
, Rockt{\"{a}}schel, T.%
\BCBL {}\ \BBA {} Riedel, S.%
\end{APACrefauthors}%
\unskip\
\newblock
\APACrefYearMonthDay{2017}{}{}.
\newblock
{\BBOQ}\APACrefatitle {Adversarial Sets for Regularising Neural Link
  Predictors} {Adversarial sets for regularising neural link
  predictors}.{\BBCQ}
\newblock
\APACjournalVolNumPages{arXiv preprint arXiv:1707.07596}{}{}{}.
\PrintBackRefs{\CurrentBib}

\bibitem [\protect \citeauthoryear {%
Nenov%
\ \protect \BOthers {.}}{%
Nenov%
\ \protect \BOthers {.}}{%
{\protect \APACyear {2015}}%
}]{%
Nenov2015}
\APACinsertmetastar {%
Nenov2015}%
\begin{APACrefauthors}%
Nenov, Y.%
, Piro, R.%
, Motik, B.%
, Horrocks, I.%
, Wu, Z.%
\BCBL {}\ \BBA {} Banerjee, J.%
\end{APACrefauthors}%
\unskip\
\newblock
\APACrefYearMonthDay{2015}{}{}.
\newblock
{\BBOQ}\APACrefatitle {{RDFox: A} Highly-Scalable {RDF} Store} {{RDFox: A}
  highly-scalable {RDF} store}.{\BBCQ}
\newblock
\BIn{} \APACrefbtitle {{Proceedings of the 14th International Semantic Web
  Conference}} {{Proceedings of the 14th International Semantic Web
  Conference}}\ (\BPGS\ 3--20).
\PrintBackRefs{\CurrentBib}

\bibitem [\protect \citeauthoryear {%
Nickel%
, Rosasco%
\BCBL {}\ \BBA {} Poggio%
}{%
Nickel%
\ \protect \BOthers {.}}{%
{\protect \APACyear {2016}}%
}]{%
Nickel2016}
\APACinsertmetastar {%
Nickel2016}%
\begin{APACrefauthors}%
Nickel, M.%
, Rosasco, L.%
\BCBL {}\ \BBA {} Poggio, T.%
\end{APACrefauthors}%
\unskip\
\newblock
\APACrefYearMonthDay{2016}{}{}.
\newblock
{\BBOQ}\APACrefatitle {Holographic Embeddings of Knowledge Graphs} {Holographic
  embeddings of knowledge graphs}.{\BBCQ}
\newblock
\BIn{} \APACrefbtitle {{Proceedings of the 30th AAAI Conference on Artificial
  Intelligence}} {{Proceedings of the 30th AAAI Conference on Artificial
  Intelligence}}\ (\BPGS\ 1955--1961).
\PrintBackRefs{\CurrentBib}

\bibitem [\protect \citeauthoryear {%
Nickel%
, Tresp%
\BCBL {}\ \BBA {} Kriegel%
}{%
Nickel%
\ \protect \BOthers {.}}{%
{\protect \APACyear {2012}}%
}]{%
Nickel2012}
\APACinsertmetastar {%
Nickel2012}%
\begin{APACrefauthors}%
Nickel, M.%
, Tresp, V.%
\BCBL {}\ \BBA {} Kriegel, H\BHBI P.%
\end{APACrefauthors}%
\unskip\
\newblock
\APACrefYearMonthDay{2012}{}{}.
\newblock
{\BBOQ}\APACrefatitle {{Factorizing YAGO: S}calable Machine Learning for
  {Linked Data}} {{Factorizing YAGO: S}calable machine learning for {Linked
  Data}}.{\BBCQ}
\newblock
\BIn{} \APACrefbtitle {{Proceedings of the 21st World Wide Web Conference}}
  {{Proceedings of the 21st World Wide Web Conference}}\ (\BPGS\ 271--280).
\PrintBackRefs{\CurrentBib}

\bibitem [\protect \citeauthoryear {%
Oaksford%
\ \BBA {} Chater%
}{%
Oaksford%
\ \BBA {} Chater%
}{%
{\protect \APACyear {2007}}%
}]{%
Oaksford2007}
\APACinsertmetastar {%
Oaksford2007}%
\begin{APACrefauthors}%
Oaksford, M.%
\BCBT {}\ \BBA {} Chater, N.%
\end{APACrefauthors}%
\unskip\
\newblock
\APACrefYear{2007}.
\newblock
\APACrefbtitle {{Bayesian Rationality: The Probabilistic Approach to Human
  Reasoning}} {{Bayesian Rationality: The Probabilistic Approach to Human
  Reasoning}}.
\newblock
\APACaddressPublisher{}{Oxford University Press}.
\PrintBackRefs{\CurrentBib}

\bibitem [\protect \citeauthoryear {%
Pollack%
}{%
Pollack%
}{%
{\protect \APACyear {1990}}%
}]{%
Pollack1990}
\APACinsertmetastar {%
Pollack1990}%
\begin{APACrefauthors}%
Pollack, J\BPBI B.%
\end{APACrefauthors}%
\unskip\
\newblock
\APACrefYearMonthDay{1990}{}{}.
\newblock
{\BBOQ}\APACrefatitle {Recursive Distributed Representations} {Recursive
  distributed representations}.{\BBCQ}
\newblock
\APACjournalVolNumPages{Artificial Intelligence}{46}{1/2}{77--105}.
\PrintBackRefs{\CurrentBib}

\bibitem [\protect \citeauthoryear {%
Rockt{\"{a}}schel%
\ \BBA {} Riedel%
}{%
Rockt{\"{a}}schel%
\ \BBA {} Riedel%
}{%
{\protect \APACyear {2017}}%
}]{%
Rocktaschel2017}
\APACinsertmetastar {%
Rocktaschel2017}%
\begin{APACrefauthors}%
Rockt{\"{a}}schel, T.%
\BCBT {}\ \BBA {} Riedel, S.%
\end{APACrefauthors}%
\unskip\
\newblock
\APACrefYearMonthDay{2017}{}{}.
\newblock
{\BBOQ}\APACrefatitle {End-to-End Differentiable Proving} {End-to-end
  differentiable proving}.{\BBCQ}
\newblock
\BIn{} \APACrefbtitle {{Advances in Neural Information Processing Systems 30}}
  {{Advances in Neural Information Processing Systems 30}}\ (\BPGS\
  3788--3800).
\PrintBackRefs{\CurrentBib}

\bibitem [\protect \citeauthoryear {%
Rockt{\"{a}}schel%
, Singh%
\BCBL {}\ \BBA {} Riedel%
}{%
Rockt{\"{a}}schel%
\ \protect \BOthers {.}}{%
{\protect \APACyear {2015}}%
}]{%
Rocktaschel2015}
\APACinsertmetastar {%
Rocktaschel2015}%
\begin{APACrefauthors}%
Rockt{\"{a}}schel, T.%
, Singh, S.%
\BCBL {}\ \BBA {} Riedel, S.%
\end{APACrefauthors}%
\unskip\
\newblock
\APACrefYearMonthDay{2015}{}{}.
\newblock
{\BBOQ}\APACrefatitle {Injecting logical background knowledge into embeddings
  for relation extraction} {Injecting logical background knowledge into
  embeddings for relation extraction}.{\BBCQ}
\newblock
\BIn{} \APACrefbtitle {{Proceedings of the 2015 Conference of the North
  American Chapter of the Association for Computational Linguistics: Human
  Language Technologies}} {{Proceedings of the 2015 Conference of the North
  American Chapter of the Association for Computational Linguistics: Human
  Language Technologies}}\ (\BPGS\ 1119--1129).
\PrintBackRefs{\CurrentBib}

\bibitem [\protect \citeauthoryear {%
Santoro%
\ \protect \BOthers {.}}{%
Santoro%
\ \protect \BOthers {.}}{%
{\protect \APACyear {2017}}%
}]{%
Santoro2017}
\APACinsertmetastar {%
Santoro2017}%
\begin{APACrefauthors}%
Santoro, A.%
, Raposo, D.%
, Barrett, D\BPBI G.%
, Malinowski, M.%
, Pascanu, R.%
, Battaglia, P.%
\BCBL {}\ \BBA {} Lillicrap, T.%
\end{APACrefauthors}%
\unskip\
\newblock
\APACrefYearMonthDay{2017}{}{}.
\newblock
{\BBOQ}\APACrefatitle {A simple neural network module for relational reasoning}
  {A simple neural network module for relational reasoning}.{\BBCQ}
\newblock
\BIn{} \APACrefbtitle {Advances in Neural Information Processing Systems 30.}
  {Advances in neural information processing systems 30.}
\PrintBackRefs{\CurrentBib}

\bibitem [\protect \citeauthoryear {%
Serafini%
\ \BBA {} d'Avila Garcez%
}{%
Serafini%
\ \BBA {} d'Avila Garcez%
}{%
{\protect \APACyear {2016}}%
}]{%
Serafini2016}
\APACinsertmetastar {%
Serafini2016}%
\begin{APACrefauthors}%
Serafini, L.%
\BCBT {}\ \BBA {} d'Avila Garcez, A.%
\end{APACrefauthors}%
\unskip\
\newblock
\APACrefYearMonthDay{2016}{}{}.
\newblock
{\BBOQ}\APACrefatitle {Logic Tensor Networks: {D}eep Learning and Logical
  Reasoning from Data and Knowledge} {Logic tensor networks: {D}eep learning
  and logical reasoning from data and knowledge}.{\BBCQ}
\newblock
\APACjournalVolNumPages{arXiv preprint arXiv:1606.04422}{}{}{}.
\PrintBackRefs{\CurrentBib}

\bibitem [\protect \citeauthoryear {%
Shastri%
}{%
Shastri%
}{%
{\protect \APACyear {1992}}%
}]{%
Shastri1992}
\APACinsertmetastar {%
Shastri1992}%
\begin{APACrefauthors}%
Shastri, L.%
\end{APACrefauthors}%
\unskip\
\newblock
\APACrefYearMonthDay{1992}{}{}.
\newblock
{\BBOQ}\APACrefatitle {{Neurally motivated constraints on the working memory
  capacity of a production system for parallel processing: Implications of a
  connectionist model based on temporal synchrony}} {{Neurally motivated
  constraints on the working memory capacity of a production system for
  parallel processing: Implications of a connectionist model based on temporal
  synchrony}}.{\BBCQ}
\newblock
\BIn{} \APACrefbtitle {{Proceedings of the 14th Annual Conference of the
  Cognitive Science Society}} {{Proceedings of the 14th Annual Conference of
  the Cognitive Science Society}}\ (\BVOL~29, \BPGS\ 159--164).
\PrintBackRefs{\CurrentBib}

\bibitem [\protect \citeauthoryear {%
Shavlik%
\ \BBA {} Towell%
}{%
Shavlik%
\ \BBA {} Towell%
}{%
{\protect \APACyear {1991}}%
}]{%
Shavlik1989}
\APACinsertmetastar {%
Shavlik1989}%
\begin{APACrefauthors}%
Shavlik, J\BPBI W.%
\BCBT {}\ \BBA {} Towell, G\BPBI G.%
\end{APACrefauthors}%
\unskip\
\newblock
\APACrefYearMonthDay{1991}{}{}.
\newblock
{\BBOQ}\APACrefatitle {An Approach to Combining Explanation-based and Neural
  Learning Algorithms} {An approach to combining explanation-based and neural
  learning algorithms}.{\BBCQ}
\newblock
\BIn{} \APACrefbtitle {{Applications of Learning and Planning Methods}}
  {{Applications of Learning and Planning Methods}}\ (\BPGS\ 231--253).
\newblock
\APACaddressPublisher{}{World Scientific}.
\PrintBackRefs{\CurrentBib}

\bibitem [\protect \citeauthoryear {%
Socher%
, Chen%
, Manning%
\BCBL {}\ \BBA {} Ng%
}{%
Socher%
\ \protect \BOthers {.}}{%
{\protect \APACyear {2013}}%
}]{%
Socher2013}
\APACinsertmetastar {%
Socher2013}%
\begin{APACrefauthors}%
Socher, R.%
, Chen, D.%
, Manning, C\BPBI D.%
\BCBL {}\ \BBA {} Ng, A\BPBI Y.%
\end{APACrefauthors}%
\unskip\
\newblock
\APACrefYearMonthDay{2013}{}{}.
\newblock
{\BBOQ}\APACrefatitle {Reasoning With Neural Tensor Networks for Knowledge Base
  Completion} {Reasoning with neural tensor networks for knowledge base
  completion}.{\BBCQ}
\newblock
\BIn{} \APACrefbtitle {{Advances in Neural Information Processing Systems 26}}
  {{Advances in Neural Information Processing Systems 26}}\ (\BPGS\ 926--934).
\PrintBackRefs{\CurrentBib}

\bibitem [\protect \citeauthoryear {%
Sourek%
, Aschenbrenner%
, Zelezny%
\BCBL {}\ \BBA {} Kuzelka%
}{%
Sourek%
\ \protect \BOthers {.}}{%
{\protect \APACyear {2015}}%
}]{%
Sourek2015}
\APACinsertmetastar {%
Sourek2015}%
\begin{APACrefauthors}%
Sourek, G.%
, Aschenbrenner, V.%
, Zelezny, F.%
\BCBL {}\ \BBA {} Kuzelka, O.%
\end{APACrefauthors}%
\unskip\
\newblock
\APACrefYearMonthDay{2015}{}{}.
\newblock
{\BBOQ}\APACrefatitle {Lifted Relational Neural Networks} {Lifted relational
  neural networks}.{\BBCQ}
\newblock
\APACjournalVolNumPages{arXiv preprint arXiv:1508.05128}{}{}{}.
\PrintBackRefs{\CurrentBib}

\bibitem [\protect \citeauthoryear {%
Sukhbaatar%
, Szlam%
, Weston%
\BCBL {}\ \BBA {} Fergus%
}{%
Sukhbaatar%
\ \protect \BOthers {.}}{%
{\protect \APACyear {2015}}%
}]{%
Sukhbaatar2015}
\APACinsertmetastar {%
Sukhbaatar2015}%
\begin{APACrefauthors}%
Sukhbaatar, S.%
, Szlam, A.%
, Weston, J.%
\BCBL {}\ \BBA {} Fergus, R.%
\end{APACrefauthors}%
\unskip\
\newblock
\APACrefYearMonthDay{2015}{}{}.
\newblock
{\BBOQ}\APACrefatitle {End-To-End Memory Networks} {End-to-end memory
  networks}.{\BBCQ}
\newblock
\BIn{} \APACrefbtitle {{Advances in Neural Information Processing Systems 28}}
  {{Advances in Neural Information Processing Systems 28}}\ (\BPGS\
  2440--2448).
\PrintBackRefs{\CurrentBib}

\bibitem [\protect \citeauthoryear {%
Sun%
\ \BBA {} Alexandre%
}{%
Sun%
\ \BBA {} Alexandre%
}{%
{\protect \APACyear {2013}}%
}]{%
Sun1997}
\APACinsertmetastar {%
Sun1997}%
\begin{APACrefauthors}%
Sun, R.%
\BCBT {}\ \BBA {} Alexandre, F.%
\end{APACrefauthors}%
\unskip\
\newblock
\APACrefYear{2013}.
\newblock
\APACrefbtitle {{Connectionist-Symbolic Integration: From Unified to Hybrid
  Approaches}} {{Connectionist-Symbolic Integration: From Unified to Hybrid
  Approaches}}.
\newblock
\APACaddressPublisher{}{Psychology Press}.
\PrintBackRefs{\CurrentBib}

\bibitem [\protect \citeauthoryear {%
Towell%
\ \BBA {} Shavlik%
}{%
Towell%
\ \BBA {} Shavlik%
}{%
{\protect \APACyear {1994}}%
}]{%
Towell1994}
\APACinsertmetastar {%
Towell1994}%
\begin{APACrefauthors}%
Towell, G\BPBI G.%
\BCBT {}\ \BBA {} Shavlik, J\BPBI W.%
\end{APACrefauthors}%
\unskip\
\newblock
\APACrefYearMonthDay{1994}{}{}.
\newblock
{\BBOQ}\APACrefatitle {{Knowledge-based artificial neural networks}}
  {{Knowledge-based artificial neural networks}}.{\BBCQ}
\newblock
\APACjournalVolNumPages{Artificial Intelligence}{70}{1/2}{119--165}.
\PrintBackRefs{\CurrentBib}

\bibitem [\protect \citeauthoryear {%
Trouillon%
\ \protect \BOthers {.}}{%
Trouillon%
\ \protect \BOthers {.}}{%
{\protect \APACyear {2017}}%
}]{%
Trouillon2017}
\APACinsertmetastar {%
Trouillon2017}%
\begin{APACrefauthors}%
Trouillon, T.%
, Dance, C\BPBI R.%
, Gaussier, {\'E}.%
, Welbl, J.%
, Riedel, S.%
\BCBL {}\ \BBA {} Bouchard, G.%
\end{APACrefauthors}%
\unskip\
\newblock
\APACrefYearMonthDay{2017}{}{}.
\newblock
{\BBOQ}\APACrefatitle {Knowledge Graph Completion via Complex Tensor
  Factorization} {Knowledge graph completion via complex tensor
  factorization}.{\BBCQ}
\newblock
\APACjournalVolNumPages{Journal of Machine Learning Research}{18}{130}{1--38}.
\PrintBackRefs{\CurrentBib}

\bibitem [\protect \citeauthoryear {%
Vendrov%
, Kiros%
, Fidler%
\BCBL {}\ \BBA {} Urtasun%
}{%
Vendrov%
\ \protect \BOthers {.}}{%
{\protect \APACyear {2016}}%
}]{%
Vendrov2016}
\APACinsertmetastar {%
Vendrov2016}%
\begin{APACrefauthors}%
Vendrov, I.%
, Kiros, R.%
, Fidler, S.%
\BCBL {}\ \BBA {} Urtasun, R.%
\end{APACrefauthors}%
\unskip\
\newblock
\APACrefYearMonthDay{2016}{}{}.
\newblock
{\BBOQ}\APACrefatitle {Order-Embeddings of Images and Language}
  {Order-embeddings of images and language}.{\BBCQ}
\newblock
\BIn{} \APACrefbtitle {{Proceedings of the 4th International Conference on
  Learning Representations}.} {{Proceedings of the 4th International Conference
  on Learning Representations}.}
\PrintBackRefs{\CurrentBib}

\bibitem [\protect \citeauthoryear {%
{Wang}%
, {Mao}%
, {Wang}%
\BCBL {}\ \BBA {} {Guo}%
}{%
{Wang}%
\ \protect \BOthers {.}}{%
{\protect \APACyear {2017}}%
}]{%
Wang2017}
\APACinsertmetastar {%
Wang2017}%
\begin{APACrefauthors}%
{Wang}, Q.%
, {Mao}, Z.%
, {Wang}, B.%
\BCBL {}\ \BBA {} {Guo}, L.%
\end{APACrefauthors}%
\unskip\
\newblock
\APACrefYearMonthDay{2017}{}{}.
\newblock
{\BBOQ}\APACrefatitle {Knowledge Graph Embedding: {A} Survey of Approaches and
  Applications} {Knowledge graph embedding: {A} survey of approaches and
  applications}.{\BBCQ}
\newblock
\APACjournalVolNumPages{IEEE Transactions on Knowledge and Data
  Engineering}{29}{12}{2724--2743}.
\PrintBackRefs{\CurrentBib}

\bibitem [\protect \citeauthoryear {%
Weston%
, Bordes%
, Chopra%
\BCBL {}\ \BBA {} Mikolov%
}{%
Weston%
, Bordes%
\BCBL {}\ \protect \BOthers {.}}{%
{\protect \APACyear {2015}}%
}]{%
Weston2015a}
\APACinsertmetastar {%
Weston2015a}%
\begin{APACrefauthors}%
Weston, J.%
, Bordes, A.%
, Chopra, S.%
\BCBL {}\ \BBA {} Mikolov, T.%
\end{APACrefauthors}%
\unskip\
\newblock
\APACrefYearMonthDay{2015}{}{}.
\newblock
{\BBOQ}\APACrefatitle {Towards {AI}-Complete Question Answering: {A} Set of
  Prerequisite Toy Tasks} {Towards {AI}-complete question answering: {A} set of
  prerequisite toy tasks}.{\BBCQ}
\newblock
\APACjournalVolNumPages{arXiv preprint arXiv:1502.05698}{}{}{}.
\PrintBackRefs{\CurrentBib}

\bibitem [\protect \citeauthoryear {%
Weston%
, Chopra%
\BCBL {}\ \BBA {} Bordes%
}{%
Weston%
, Chopra%
\BCBL {}\ \BBA {} Bordes%
}{%
{\protect \APACyear {2015}}%
}]{%
Weston2015}
\APACinsertmetastar {%
Weston2015}%
\begin{APACrefauthors}%
Weston, J.%
, Chopra, S.%
\BCBL {}\ \BBA {} Bordes, A.%
\end{APACrefauthors}%
\unskip\
\newblock
\APACrefYearMonthDay{2015}{}{}.
\newblock
{\BBOQ}\APACrefatitle {Memory Networks} {Memory networks}.{\BBCQ}
\newblock
\BIn{} \APACrefbtitle {{Proceedings of the 3rd International Conference on
  Learning Representations}.} {{Proceedings of the 3rd International Conference
  on Learning Representations}.}
\PrintBackRefs{\CurrentBib}

\bibitem [\protect \citeauthoryear {%
Xu%
, Zhang%
, Friedman%
, Liang%
\BCBL {}\ \BBA {} {Van den Broeck}%
}{%
Xu%
\ \protect \BOthers {.}}{%
{\protect \APACyear {2018}}%
}]{%
Xu2018}
\APACinsertmetastar {%
Xu2018}%
\begin{APACrefauthors}%
Xu, J.%
, Zhang, Z.%
, Friedman, T.%
, Liang, Y.%
\BCBL {}\ \BBA {} {Van den Broeck}, G.%
\end{APACrefauthors}%
\unskip\
\newblock
\APACrefYearMonthDay{2018}{}{}.
\newblock
{\BBOQ}\APACrefatitle {A Semantic Loss Function for Deep Learning with Symbolic
  Knowledge} {A semantic loss function for deep learning with symbolic
  knowledge}.{\BBCQ}
\newblock
\BIn{} \APACrefbtitle {{Proceedings of the 35th International Conference on
  Machine Learning}} {{Proceedings of the 35th International Conference on
  Machine Learning}}\ (\BPGS\ 5498--5507).
\PrintBackRefs{\CurrentBib}

\end{thebibliography}
\bibliographystyle{apacite}


\begin{acronym}
    \acro{SGD}{stochastic gradient descent}
\end{acronym}

\end{document}